\documentclass{article}
\usepackage{ijcai25}

\usepackage[table,xcdraw]{xcolor}
\usepackage{times}
\usepackage{soul}
\usepackage{url}
\usepackage[hidelinks]{hyperref}
\usepackage{marvosym}
\usepackage[utf8]{inputenc}
\usepackage[small]{caption}
\usepackage{graphicx}
\usepackage{caption}
\usepackage{amsmath}
\usepackage{amsthm}
\usepackage{booktabs}
\usepackage{algorithm}
\usepackage[switch]{lineno}
\usepackage{verbatim}
\usepackage{enumitem}

\usepackage{latexsym}
\usepackage{amssymb}
\usepackage[capitalize]{cleveref}
\usepackage{multirow}
\usepackage{algpseudocode}
\usepackage{appendix}
\usepackage{float}
\usepackage{newfloat}
\usepackage{tabularx} 
\usepackage{ragged2e} 
\usepackage{dsfont}
\newcolumntype{L}{>{\raggedright\arraybackslash}X}
\newcolumntype{C}{>{\centering\arraybackslash}X}
\newcolumntype{R}{>{\raggedleft\arraybackslash}X}
\usepackage{color}
\definecolor{plum}{RGB}{142,69,133}
\definecolor{brown}{RGB}{255,222,173}
\definecolor{green}{RGB}{143,188,143}
\definecolor{blue}{RGB}{70,130,180}

\title{
    The Devil is in Fine-tuning and Long-tailed Problems: 
    \\ A New Benchmark for Scene Text Detection
}

\author{
Tianjiao Cao$^{1,3,*}$
\and
Jiahao Lyu$^{1,3,*}$
\and
Weichao Zeng$^{1,3}$
\and
Weimin Mu$^{1}$
\And
Yu Zhou$^{2,}$\textsuperscript{\Letter}
\affiliations
$^1$Institute of Information Engineering, Chinese Academy of Sciences\\
$^2$VCIP \& TMCC \& DISSec, College of Computer Science, Nankai University \\
$^3$School of Cyber Security, University of Chinese Academy of Sciences \\
\emails
\{caotianjiao, lvjiahao, zengweichao, muweimin\}@iie.ac.cn, 
yzhou@nankai.edu.cn
}

\begin{document}

\maketitle

\begin{abstract}
Scene text detection has seen the emergence of high-performing methods that excel on academic benchmarks. However, these detectors often fail to replicate such success in real-world scenarios. We uncover two key factors contributing to this discrepancy through extensive experiments.
First, a \textit{Fine-tuning Gap}, where models leverage \textit{Dataset-Specific Optimization} (DSO) paradigm for one domain at the cost of reduced effectiveness in others, leads to inflated performances on academic benchmarks. 
Second, the suboptimal performance in practical settings is primarily attributed to the long-tailed distribution of texts, where detectors struggle with rare and complex categories as artistic or overlapped text.
Given that the DSO paradigm might undermine the generalization ability of models, we advocate for a \textit{Joint-Dataset Learning} (JDL) protocol to alleviate the Fine-tuning Gap. 
Additionally, an error analysis is conducted to identify three major categories and 13 subcategories of challenges in long-tailed scene text, upon which we propose a Long-Tailed Benchmark (LTB).
LTB facilitates a comprehensive evaluation of ability to handle a diverse range of long-tailed challenges. We further introduce MAEDet, a self-supervised learning-based method, as a strong baseline for LTB.
The code is available at \url{https://github.com/pd162/LTB}.
\end{abstract}

\begin{figure}[!t]
\centering\includegraphics[width=0.48\textwidth]{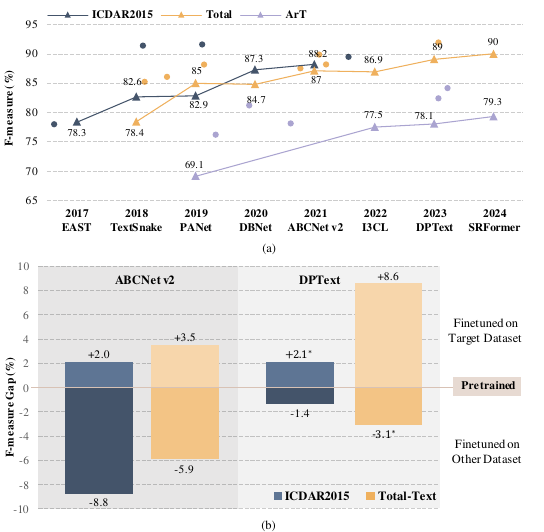}
    \caption{(a): The performance tendency of scene text detectors in recent years. We choose several representative methods in chronological order. Circles indicate the performances of other methods.  (b): A visualization of fine-tuning gap. Each bar shows a performance comparison on the same benchmark. ``$*$'’ denotes we fine-tuned the model with the open-source pretrained weight. 
    }
    \label{fig:tendency}
\end{figure}

\section{Introduction}

\let\thefootnote\relax
\footnotetext{{\normalsize * \ Equal Contribution.} {\normalsize \Letter \ Corresponding Author.} }

Scene Text Detection (STD) has received extensive attention both in academia and industry due to its wide applications, such as scene understanding \cite{zhang2025linguistics,lyu2025arbitrary}, 
intelligent office \cite{zeng2023filling,shen2024falcon,shen2023divide}, 
etc. 
Researchers have proposed many prominent scene text detectors~\cite{zhou2017east,long2018textsnake,wang2019efficient,liao2020real,liu2021abcnet,du2022i3cl,qin2023towards,bu2024srformer}. %
\Cref{fig:tendency} (a) shows the year-over-year performance improvement of scene text detectors across three public benchmarks. Recent advances in scene text detection show consistent improvements in F-measure scores, with recent models surpassing 90\%, reflecting rapid progress in the field.

Despite the impressive benchmark results, scene text detectors still face challenges (e.g., texts within bad illumination and complex backgrounds) and fail to meet the expectations in real-world applications.
The significant discrepancy between the high performance on academic benchmarks and the subpar performance in practical deployments prompts our investigation into the underlying causes. We attribute this to two main factors: fine-tuning gap and long-tailed problem.

Firstly, previous evaluations of scene text detectors mainly rely on the \textit{Dataset-Specific Optimization} (DSO) paradigm, where detectors are pretrained on a variety of datasets and then fine-tuned on a single dataset's training subset, followed by assessment on its corresponding test subset. However, such a protocol might not reflect authentic generalization ability, as each dataset exhibits unique biases and domain-specific characteristics. Fine-tuning on a biased training subset could lead to overfitting to a specific domain, resulting in degraded performance on others. 
As shown in \Cref{fig:tendency}(b), a model consistently improves its performance on the test set after fine-tuning with the corresponding training data. Conversely, its performance drops significantly, with a decrease of up to 8.8\% when fine-tuned on one dataset and evaluated on another.
Evaluations on limited benchmarks reflect the detector's effectiveness for specific text types but fail to assess its overall performance in diverse real-world scenarios. Previous works \cite{zhan2019ga,tian2022domain} also refer to the domain-adaptation problem in scene text detection. 
To address this issue, we propose a \textit{Joint-Dataset Learning} (JDL) method, where a detector is trained on a unified training set from multiple datasets and directly evaluated on a combined test set. We posit that JDL offers a more accurate and comprehensive performance assessment.

\begin{figure}[t]
\centering
\includegraphics[width=0.48\textwidth]{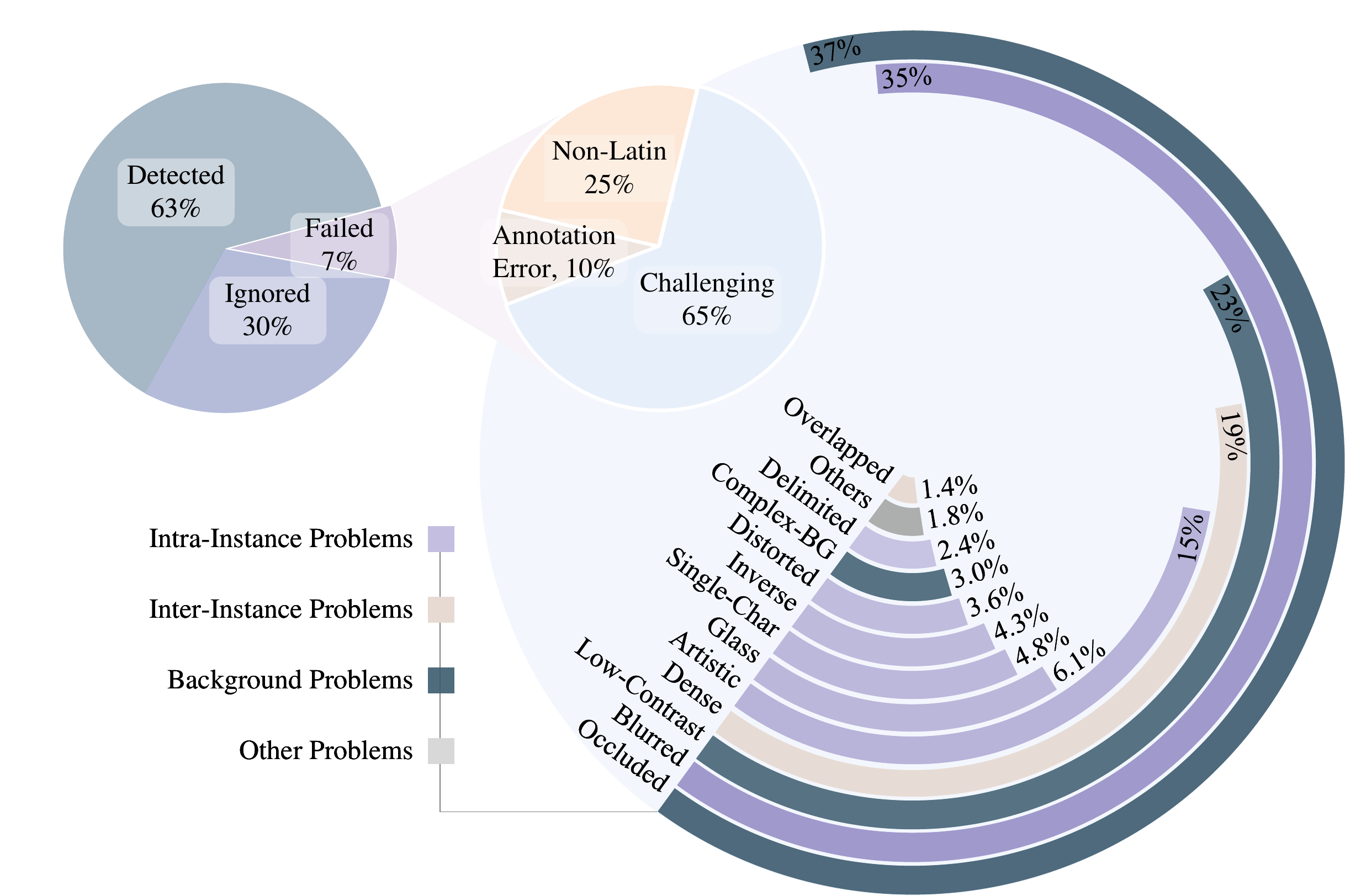}
    \caption{
        A quantitative analysis of failure cases. The derivation of results is described in \Cref{sec:benchmark}.  The left figure shows that the overall detection results are divided into ignored, detected, and failed types. The middle presents statistics of failure cases with human annotation. The right illustrates the distribution of challenging text instances.
        The brightness of color indicates the prevalence of each category, with darker colors representing higher proportions. BG for background and Char for character.
    }
    \label{fig:problem}
\end{figure}

Secondly, scene text categories commonly exhibit a long-tailed distribution in real-world scenarios, with dominant head categories (e.g., printed text) and scarce tail categories (e.g., artistic text).  We define long-tailed problems as challenges where detectors struggle with tail categories due to the lack of relevant training samples, resulting in substantial performance degradation on less frequent but critical text instances in practical applications.
However, most existing studies focus on specific challenges and overlook long-tailed issues from a holistic perspective. In light of this, we analyze and define a comprehensive set of challenges, as shown in \Cref{fig:problem}. We evaluate four representative scene text detectors (CNN-based and Transformer-based) on three datasets with distinct styles (for diverse scenarios). Results reveal that the detectors fail to identify 7\% of text instances. Among these undetected instances, only 65\% are found to be genuinely challenging, while the remaining 35\% are attributed to annotation errors or involved non-Latin scripts. We identify recurring patterns across these challenging cases and categorize them into three main groups: intra-instance, inter-instance, and background problems. These groups are further divided into 13 subcategories, each representing a long-tailed challenge in scene text detection. To support future research, we introduce a Long-Tailed Benchmark (LTB) as a standardized platform for evaluating the ability of scene text detectors to address these challenges.

Finally, inspired by MAERec \cite{jiang2023revisiting}, we utilize self-supervised learning (SSL) for better text representation and propose a baseline based on MAE \cite{he2022masked} from the data perspective, called MAEDet. 
In conclusion, our contributions are three-fold.

\begin{itemize}
    \item We identify the devil of the discrepancy between the performance on academic benchmarks and that on real applications lies in fine-tuning and long-tailed problems.
    
    \item We introduce the \textit{Joint-Dataset Learning} (JDL) protocol as a replacement for \textit{Dataset-Specific Optimization} to avoids the fine-tuning gap and provides a more accurate evaluation of model generalization across different domains.
    
    \item We propose the Long-Tailed Benchmark (LTB), which poses 13 distinct challenges and serves as the first benchmark to comprehensively evaluate models' capability to address various long-tailed problems in scene text detection. A baseline based on self-supervised learning for the benchmark is also provided.
\end{itemize}

\section{Related Works}

\subsection{Approaches for Scene Text Detection}

\textbf{CNN-based methods} \cite{gupta2016synthetic,jaderberg2016reading,liao2017textboxes,liao2018textboxes++} are followed the object detection method to detect the scene text. \cite{he2017mask} and its variants  \cite{xiao2020sequential,du2022i3cl,ye2020textfuse} represent the oriented and curve text contours.
\cite{long2018textsnake,liu2020abcnet,wang2020textray,wang2022tpsnet,liu2023pbformer,su2024lranet} also explore the efficient representation operators. 

\textbf{Transformer-based methods}
Attributing to the broad application in visual tasks, Transformer, as the fundamental architecture, has been gradually embedded in the scene text detector \cite{raisi2021transformer,tang2022few,liu2023pbformer,bu2024srformer}. \cite{raisi2021transformer} firstly introduces the Deformable DETR \cite{zhu2020deformable} to detect oriented text instances. 
To alleviate the burden of the Transformer, \cite{tang2022few} samples the features of text instances and reduces the computational cost. \cite{ye2023dptext} notices the reading of inverse-like scene text. \cite{bu2024srformer} combines the advantages of top-down and bottom-up and proposes a transformer-powered text detector. 

\subsection{Benchmarks for Scene Text Detection}
\label{para:benchmark}
Thanks to the Robust Reading Contest, massive and diversified datasets are proposed and evaluated fairly, such as ICDAR 2013 \cite{karatzas2013icdar} for axis-align texts, ICDAR 2015 \cite{karatzas2015icdar} for oriented and unfocused texts, ArT \cite{chng2019icdar2019} for arbitrarily shaped texts, and MLT \cite{nayef2017icdar2017,nayef2019icdar2019} for multi-language texts. These datasets include thousands of text images for training and testing. LSVT \cite{sun2019icdar} and TextOCR \cite{singh2021textocr} collect larger-scale annotated datasets. Additionally, some long-tailed problems in scene text detection have been noticed in recent years. For example, SCUT-CTW1500 \cite{yuliang2017detecting} is designed to handle curved text with line-level granularity. DAST1500 \cite{tang2019seglink++}, HierText \cite{long2022towards}, and InverseText \cite{ye2023dptext} solve the problems in dense, multi-hierarchical and inversed texts, respectively.
Our current research focuses on English word-level texts, with prospects for future expansion to additional languages and varied granularity.

\section{Addressing the Fine-Tuning Gap}
\label{sec:gap}

\begin{figure}[!t]
\centering
\includegraphics[width=0.48\textwidth]{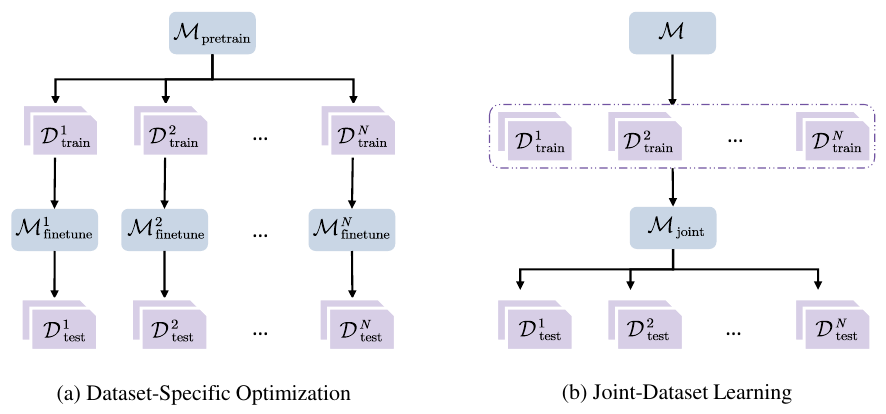}
    \caption{
        Comparison of the Dataset-Specific Optimization (DSO) and Joint-Dataset Learning (JDL) paradigms.
    }
    \label{fig:paradigm}
\end{figure}

\begin{table}[t]
\centering
\small
\begin{tabularx}{\linewidth}{@{}
        p{1.9cm}
        *{4}{>{\centering\arraybackslash}p{0.6cm}}
        CC
        @{\hspace{0cm}}X
    }
      \toprule[1pt]
      \multirow{2}{*}{Method} & 
      \multicolumn{2}{c}{IC15} &
      \multicolumn{2}{c}{TT} &
      \multicolumn{1}{c}{IC15} &
      \multicolumn{1}{c}{TT $\rightarrow$} \\
      \cmidrule(lr){2-5} 
      \multicolumn{1}{c}{\multirow{1}{*}{}} & 
      \multicolumn{1}{c}{\multirow{1}{*}{P}} & 
      \multicolumn{1}{c}{\multirow{1}{*}{F}} & 
      \multicolumn{1}{c}{\multirow{1}{*}{P}} & 
      \multicolumn{1}{c}{\multirow{1}{*}{F}} &
      \multicolumn{1}{c}{$\rightarrow$ TT} & 
      \multicolumn{1}{c}{IC15}\\

      \midrule[1pt]
      ABCNet v2 & 86.2$^\dagger$ & 88.2$^\dagger$ & 83.7$^\dagger$ & 87.2$^\dagger$ & 77.8 & 77.4 \\
      DPText-DETR & 75.3 & 77.4 & 80.4 & 89.0$^\dagger$ & 77.3 & 73.9 \\
      
      \bottomrule[1pt]
\end{tabularx}

\caption{
Statistics for fine-tuning gap. F-measure (\%) is reported with ``P'' for the pretrained model and ``F'' for the fine-tuned model. ``{A} $\rightarrow$ {B}'' denotes fine-tuning on the training set of {A} and evaluation on test set of {B}. ``$\dagger$'' denotes the official results from the corresponding paper.
}
\label{tab:finetuning_gap}
\end{table}

Traditional scene text detection methods often adopt the DSO paradigm, which can be formulated as follows.
Given $N$ datasets $\mathcal{D} = \{ \mathcal{D}^{i}\mid i=1,2,\cdots, N \}$, where each dataset $\mathcal{D}^{i}$ consists of a training set $\mathcal{D}_{\text{train}}^{i}$ and a testing set $\mathcal{D}_{\text{test}}^{i}$, a pretrained model $\mathcal{M}_{\text{pretrain}}$ is first obtained by training on large-scale data (typically synthetic data or a combination of multiple real datasets). The pretrained model is then fine-tuned separately on each training set $\mathcal{D}_{\text{train}}^{i}$ to produce $N$ fine-tuned models $\mathcal{M}_{\text{finetune}}^i$, which are finally evaluated on their corresponding test sets $\mathcal{D}_{\text{test}}^{i}$, as illustrated in \Cref{fig:paradigm} (a).

However, different benchmarks are tailored to solve various issues (e.g., oriented text detection, curved text detection) and exhibit significant variability due to diverse scenarios and data collection methods, leading to substantial dataset biases. A comprehensive analysis of these variations is provided in Appendix.
This issue is further exacerbated by the DSO protocol, where fine-tuning a model on a single training set $ \mathcal{D}_{\text{train}}^i$ might not adequately capture the overall distribution of real-world scenarios, leading to performance degradation on other test sets $\mathcal{D}_{\text{test}}^{ \setminus i}$. 

We define this limitation as the \textit{Fine-tuning Gap}: the discrepancy between a model’s performance on its fine-tuned dataset and its ability to generalize to unseen data distributions. 
To verify the fine-tuning gap, we conduct experiments using CNN-based ABCNet v2\footnote{https://github.com/Yuliang-Liu/bezier\_curve\_text\_spotting} and transformer-based DPText-DETR\footnote{https://github.com/ymy-k/DPText-DETR}, with the results presented in \Cref{tab:finetuning_gap}. As shown in the first two columns, fine-tuning yields at least a 2.0\% improvement on the target benchmark. However, the last two columns reveal degraded performance when models are fine-tuned on another dataset, which occurs irrespective of the detector architecture, whether CNN-based or Transformer-based.
These findings underscore that the DSO approach is ill-suited for actual applications due to several critical issues: 1) Lack of generalization. Detectors fine-tuned on a single dataset struggle to adapt to real-world scenarios' diverse and variable conditions, often resulting in overfitting. 2) Resource inefficiency. Separate fine-tuning for each dataset requires substantial computational and time resources, significantly limiting scalability and practicality. 3) Unrepresentative benchmarking: Evaluating a model on a single, limited test set fails to provide an objective and comprehensive assessment, potentially leading to misleading conclusions.

To address the aforementioned issues, we introduce \textit{Joint-Dataset Learning} (JDL) protocol to STD, as illustrated in \Cref{fig:paradigm} (b). JDL draws inspiration from previous advancements in scene text recognition and offers a feasible solution from data perspective. The protocol involves training a model $ \mathcal{M}_{\text{joint}} $ on the combined diversity of multiple datasets $\bigcup_{i=1}^N \mathcal{D}_{\text{train}}^i$, and directly evaluating it on $\bigcup_{i=1}^N \mathcal{D}_{\text{test}}^i$. This unified paradigm offers a more efficient training framework and a more equitable assessment mechanism for STD in practical applications.

We suggest that this new protocol provides a more comprehensive and accurate reflection of a model's adaptability to the complexities of real-world scenarios. To provide a more comprehensive view of the advantages and implications of Joint-Dataset Learning, we delve deeper into its importance and feasibility in Appendix.

\section{A Long-tailed Benchmark}\label{sec:benchmark}
\begin{figure*}[t]
  \centering
  \includegraphics[scale=0.3]{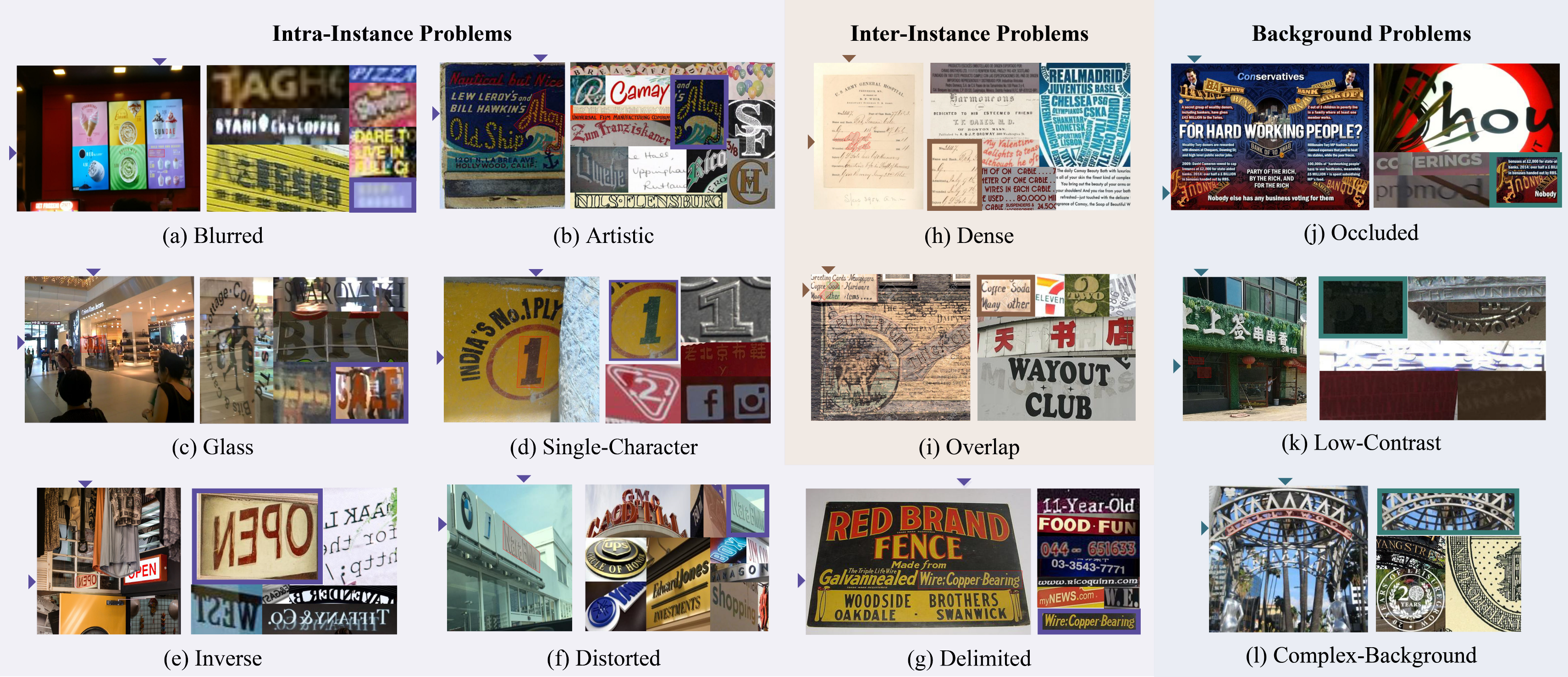}
  \caption{
  An overview of LTB which collects the challenging issues in scene text detection, involving the \textit{Intra-Instance Problems}, \textit{Inter-Instance Problems}, and \textit{Background Problems}.
  Except for the \textit{others} category, there are 12 kinds of challenges in total, each is illustrated with two images. Left: a complete scene featuring challenging text instances, highlighted by red masks and arrows pointing to their locations.
  Right: a series of cropped challenging text instances, with the corresponding instance from the left image highlighted by a bold border.
  }
  \label{fig:challenge_definition}
\end{figure*}

Except for the fine-tuning gap, scene text detection also suffers from the problem of long-tailed distribution. Text instances of tailed categories rarely exist in the training set, bringing obstacles to vanilla text detectors. In this section, we will define the common problems of current detection systems, and introduce our approach to design a benchmark to address them.

\subsection{Challenge Definition}
To uncover the challenges encountered in scene text detection, we first apply the four representative text detectors mentioned in \Cref{tab:data_collection} to the test sets of ICDAR2015 \cite{karatzas2015icdar}, Total-Text \cite{ch2017total}, and ArT \cite{chng2019icdar2019}. By analyzing the failure cases of these detectors, we identify recurring features among the challenging texts. Based on the comprehensive analysis, we categorize the undetected issues into three main groups: intra-instance problems, inter-instance problems, and background problems. Each category poses specific issues that impact text detection in distinct ways. \Cref{fig:challenge_definition} visualizes these categories and their relationships.

\subsubsection{Intra-instance Problems}
\label{sec:intra_instance_problems}
Intra-instance problems refer to challenges associated with the internal attributes of independent text instances. In other words, these problems derive from the low-quality appearance of texts rather than influence from adjacent texts or surrounding backgrounds.

\textbf{Blurred Text.} 
Blurred text may arise from out-of-focus conditions, like camera shake, limited camera performance, or image noise. Especially in scene images, texts are often not dominant and easy to be unfocused. As shown in \Cref{fig:challenge_definition}(a), blurred text is prevalent in scene text.

\textbf{Artistic Text.} 
Artistic texts frequently appear in graffiti, posters, and advertisements, characterized by elaborate but unconventional designs. These texts may feature highly unique fonts, diverse colors, and non-traditional character arrangements, such as staggered placements where characters are not aligned in a straight line. The non-standard instinct of artistic texts makes it difficult for detectors to generalize from training data, necessitating the development of advanced algorithms to handle these diverse forms effectively.
 
\textbf{Glass Text.}
Glass texts refer to texts that are either directly on the surface of glass or reflected onto it. The optical properties of glass, including reflection, refraction, and distortion, present unique challenges for text detection.

\textbf{Single-Character Text.} 
In practical applications, individual characters may be large and clear yet still hard to detect because the lack of surrounding contextual clues from adjacent characters or words prevents the detection process. 

\textbf{Distorted Text.} 
Text distortion may arise from several factors: capturing text from highly oblique angles, applying perspective transformations to the text image, or the existence of texts on irregular surfaces. Such misalignment and warping alter the original text proportions and geometric features, creating substantial challenges for detection systems.

\textbf{Inverse Text.} 
Inverse text occurs when a surface is visible from both sides, with one side showing the text normally and the other side displaying it in reverse.
It can also result from image processing techniques like horizontal flipping.  DPText-DETR\cite{ye2023dptext} argues that inverse-like text instances harm the robustness of the scene text detector.

\textbf{Delimited Text.} 
A delimited text instance contains formats where specific delimiters segment the content, resulting in multiple discrete pieces rather than a unified whole. Examples include URLs (e.g., www.example.com), phone numbers (e.g., tel:123-4567), and numerical sequences (e.g., 1,0978). The segmentation can result in detection systems losing smaller delimiters and treating the segments as separate text instances, complicating the detection and analysis of overall content.

\subsubsection{Inter-instance Problems}
In addition to the challenges posed by individual text instances, the interaction among adjacent texts also exerts an influence. Specifically, dense and overlapped texts complicate text detection processes.

\textbf{Dense Text.} Dense text refers to instances where text is closely positioned, either in multi-line or side-by-side formats. This closeness may result in the incorrect merging of separate text instances into one, which is a common issue in document-like images.

\textbf{Overlapped Text.} When text instances overlap, they become partially obstructed but remain visible. This commonly occurs due to shadows cast on the text, reflections from glossy surfaces, or the presence of watermarks.

\label{sec:background_problems}
\subsubsection{Background Problems}
The challenges associated with noisy backgrounds stem from intricate or distracting patterns and colors. When backgrounds compete with text for visual attention, it complicates the detection process.

\textbf{Occluded Text.} Occluded text occurs when text instances are partially invisible due to being obscured by foreground objects, excessive illumination, or being located at the edge of an image where parts of the text extend beyond the image boundary. This phenomenon commonly arises in natural scenes, resulting in missing segments.

\textbf{Low-Contrast Text.} Low-contrast text refers to text that has a low contrast with its background. This typically occurs when the text color is similar to the background color or when the text is inadequately illuminated. Such text is challenging for humans to detect even in high-quality images, making it even more difficult for detection algorithms to identify.

\textbf{Complex-Background Text.} 

It is difficult to discern the text boundaries when the background contains patterns that closely resemble the text or when the background features complex, multi-colored designs. For instance, text overlaid on highly textured or patterned backdrops or hollowed text that is outlined rather than filled tends to confuse text detectors.

\label{sec:other_problems}
\subsubsection{Other Problems} 
Additionally, some challenges remain difficult to categorize because we lack a clear understanding of why certain texts, though evident to human observers, elude scene text detectors. These complexity and rarity cases are grouped into this category, underscoring the need for further investigation into these atypical failures. 

\subsection{Pipeline of Constructing Benchmark}
\label{sec:ltb_pipline}
\begin{table}[!t]
\centering
\small
\begin{tabularx}{\linewidth}{@{}
    >{\raggedright\arraybackslash}p{0.1cm}|
    >{\raggedright\arraybackslash}p{2.2cm}|@{}
    C@{}
    C@{}
    C@{}
    C
}
      \toprule
      \multicolumn{1}{c}{} & 
      \multicolumn{1}{l}{Dataset} & 
      \multicolumn{1}{c}{DBNet++} & 
      \begin{tabular}{@{}c@{}}ABCNet\\v2\end{tabular} & 
      \begin{tabular}{@{}l@{}}DPText-\\DETR\end{tabular} & 
      SRFormer  \\
      
      \midrule
      \multirow{3}{*}{\rotatebox[origin=c]{90}{Detector\hspace{0cm}}}
      	& ICDAR2015 & \checkmark  & \checkmark  & & \\
        & Total-Text & \checkmark  & \checkmark  & \checkmark  & \checkmark \\
        & ArT & \checkmark  & & \checkmark  & \checkmark \\
                                   
      \midrule
      \multirow{3}{*}{\rotatebox[origin=c]{90}{Manual\hspace{0cm}}}
      	& Inverse-Text &- &  - & - & - \\
      	& NightTime-ArT &- & - & - & - \\
        & ORT & - &- &-  & - \\
      \bottomrule

\end{tabularx}
  \caption{Dual strategy of dataset collection. ``$\checkmark$'' indicates detector is used for processing the dataset. ``ORT'' refer to Occluded RoadText.
  }
  \label{tab:data_collection}
\end{table}

While these unresolved challenges have been identified, existing benchmarks either do not focus on them or only concentrate on one specific problem, limiting their utility in evaluating model performance across diverse and complex scenarios. To address this gap, we introduce the long-tailed Benchmark (LTB), which consists of 924 images, including a wide range of challenging text instances in various real-world scenes. LTB provides a more rigorous and comprehensive standard for assessing the performance of text detection models.

Our strategy for data collection prioritizes two main criteria: ensuring diversity in scenes and incorporating text instances that present predefined challenges. 
Based on these criteria, we choose six datasets, which are ICDAR2015 \cite{karatzas2015icdar}, Total-Text \cite{ch2017total}, ArT \cite{chng2019icdar2019}, InverseText \cite{ye2023dptext}, Occluded-RoadText\footnote{https://rrc.cvc.uab.es/?ch=29}, and NightTime-ArT \cite{yu2023turning}. Detailed information on these benchmarks can be found in Appendix.
According to the characteristics of different benchmarks, we adopt a dual strategy for data collection: one assisted by detectors and one purely manual.
For ICDAR2015, Total-Text, and ArT, we obtain detection results of 4 well-finetuned detectors, as outlined in \Cref{tab:data_collection}. 
We then develop an algorithm to retain undetected text instances.
For the remaining three datasets, we manually collect challenging samples due to the absence of fine-tuned models demonstrating superior performance. Appendix elaborates our data processing strategy of LTB.

\begin{figure}[!t]
\centering
\includegraphics[width=0.48\textwidth]{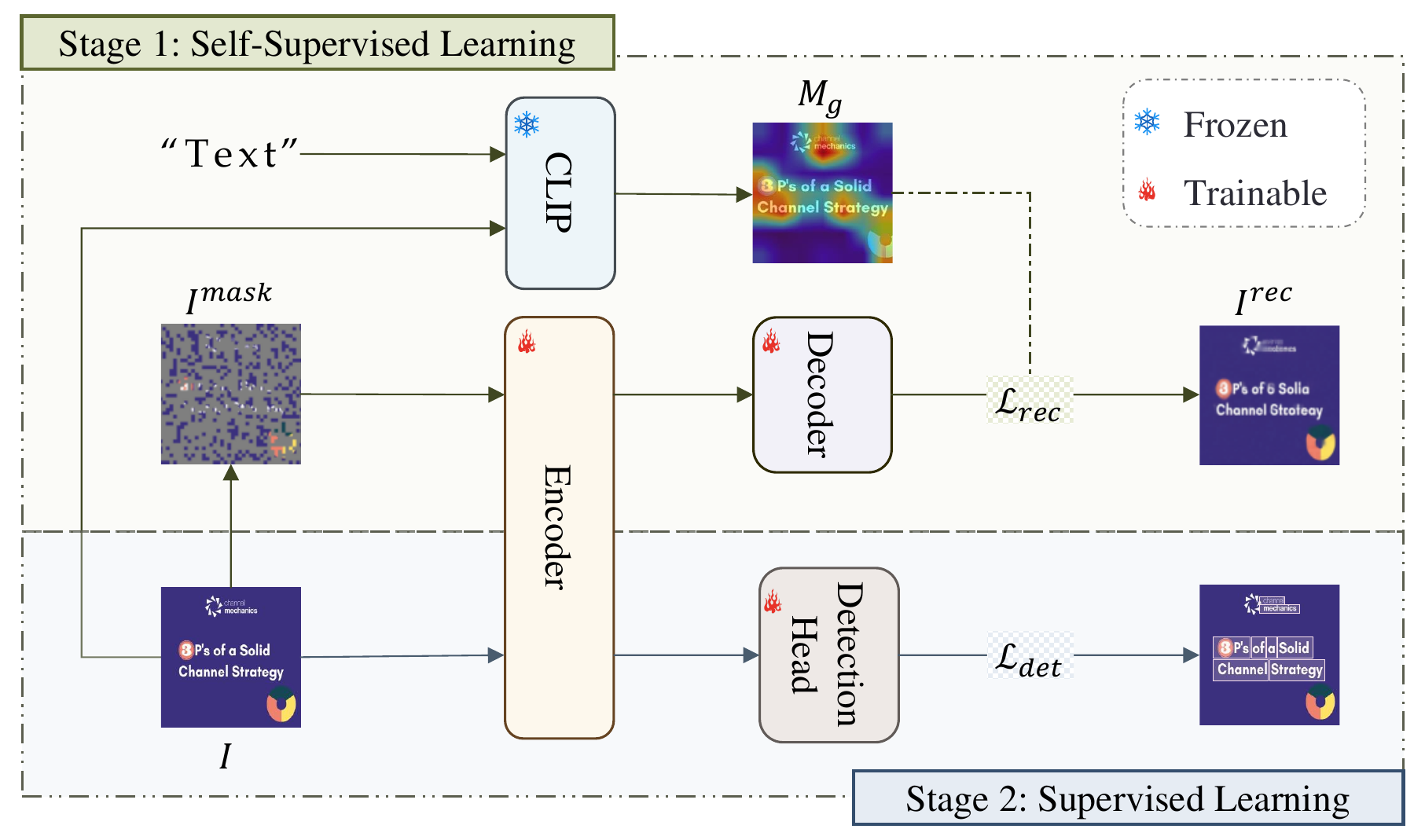}
    \caption{
    A detailed illustration of MAEDet architecture and training objectives. As indicated by its name, MAEDet is based on Mask Auto-Encoder (MAE), with the DBNet detection head adopted for text detection.
    }
    \label{fig:baseline}
\end{figure}

\begin{table*}[!tb]
  \tiny
  \centering
  \resizebox{\linewidth}{!}{
    \begin{tabular}{l|llcccccccccc}
      \toprule 
      \multicolumn{1}{c}{}& 
      \multicolumn{1}{l}{\multirow{2}{*}{Method}} & 
      \multicolumn{1}{l}{\multirow{2}{*}{Train Dataset}} &
      \multicolumn{9}{c}{Test Dataset} \\
      \cmidrule(l){4-13}\multicolumn{1}{c}{} & \multicolumn{1}{l}{} & \multicolumn{1}{l}{} &
      \begin{tabular}[c]{@{}c@{}}ICDAR\\ 2013\end{tabular} & 
      \begin{tabular}[c]{@{}c@{}}ICDAR\\ 2015\end{tabular} & 
      \begin{tabular}[c]{@{}c@{}}TT\end{tabular} & 
      \begin{tabular}[c]{@{}c@{}}COCO\end{tabular} & 
      \begin{tabular}[c]{@{}c@{}}TextOCR\end{tabular} &
      ArT & LSVT &
      \begin{tabular}[c]{@{}c@{}}MLT\\ 2017\end{tabular} & 
      \begin{tabular}[c]{@{}c@{}}MLT\\ 2019\end{tabular} &
      Avg.
      \\
      
      \midrule 
      \multirow{4}{*}{\rotatebox[origin=c]{90}{Pretrain\hspace{0pt}}} & 
      DBNet++ & ST800K & 66.6 & 22.8 & 39.5 & 27.5 & 11.1 & 26.7 & 12.1 & 56.0 & 52.3 & 35.0 \\
      & ABCNet v2 & ST150K+COCO$^*$+MLT19$^*$ & 88.1 & 78.0 & 81.0 & 53.2 & 40.5 & 69.3 & 48.2 & 87.0 & 83.6 & 69.9 \\
      & DPText & ST150K+TT+MLT19 & 89.3 & 75.3 & 80.4 & 59.5 & 43.5 & 72.1 & 52.3 & 84.2 & 83.4 & 71.1 \\ 
      & SRFormer & ST150K+TT+MLT17 &  \textbf{90.6} & \textbf{81.5} &  {84.4} & 59.4 & 51.9 & 75.0 & 59.2 &  \textbf{90.1} &  \textbf{86.8} & 75.4 \\

      \midrule 
      \multirow{6}{*}{\rotatebox[origin=c]{90}{Finetune\hspace{0pt}}} 
      & DBNet & ICDAR2015 & 75.8	& 84.0	& 69.5	& 60.5	& 61.0	& 49.9	& 16.1	& 75.9	& 76.2 & 63.2 \\
      & DRRG & CTW1500 & 42.0 & 44.0 & 39.7 & 31.9 & 18.8 & 48.0 & 56.8 & 37.5 & 41.2 & 40.0 \\
      & FCENet & Total-Text & 75.4 & 72.8 & 82.6 & 57.9 & 48.6 & 62.2 & 33.3 & 80.0 & 75.2 & 65.3 \\
      & LRANet & Total-Text & 85.4 & 79.7 & \textbf{88.9} & 61.7 & 57.3 & 66.7 & 21.6 & 86.6 & 82.4   & 70.0 \\
      & PSENet & ICDAR2015 & 60.8 & 78.7 & 53.5 & 56.0 & 50.8 & 38.7 & 13.9 & 58.6 & 57.6 & 52.0 \\
      & TCM & ICDAR2015 & 81.6 & 88.0 & 79.0 & 65.9 & 67.0 & 57.6 & 18.2 & 82.6 & 82.8 & 69.2 \\

      \midrule
      \multirow{3}{*}{\rotatebox[origin=c]{90}{Joint\hspace{0pt}}} & 
      DBNet++ & Joint98K & 87.1 & 69.3 &  {84.4} &  {59.5} & 58.3 & 78.9 & 73.4 & 83.3 & 77.1 & 74.6 \\
      & PANet & Joint98K & 82.9 & 75.6 & 75.7 & \textbf{64.5} & 54.0 & 74.8 & 72.2 & 80.0 & 79.7 & 73.3 \\
      & DPText & Joint98K & 88.2 & 75.9 & 81.9 & 57.3 &  \textbf{61.6} &  \textbf{88.2} &  \textbf{75.9} & 87.5 & 80.3 &  \textbf{77.4} \\
      \midrule
      \multirow{2}{*}
      {\rotatebox[origin=c]{90}{SSL\hspace{0pt}}}
       & MAEDet$^\dagger$  & Joint98K & 85.3 & 61.3 & 79.4 & 54.1 & 56.3 & 74.8 & 68.5 & 81.1 & 76.4 & 70.8 \\
       & MAEDet & Joint98K & 87.1 & 67.5 & 82.8 & 58.7 & 60.6 & 78.4 & 71.8 & 83.3 & 80.3 & 74.5 \\
      
      \bottomrule 
    \end{tabular}
  }
  \caption{
    Comparison of MAEDet with other STD models across all nine test sets within Joint98K. F-measure (\%) is reported.
    ``$*$'' denotes only a portion of the training images are used. ``$\dagger$'' denotes model is trained without self-supervised weights. 
    Abbreviations: TT = Total-Text, COCO = COCO-Text, ST800K = SynthText800K, ST150K = SynthText150K, SSL=self-supervised learning. \textbf{Bold} indicates the best result.
  }
  \label{tab:eval_on_9_datasets}
\end{table*}

\begin{table*}[!tb]
  \Large
  \centering
  \renewcommand{\arraystretch}{1.22}
  \resizebox{\linewidth}{!}{
    \begin{tabular}{l|lccccccc|cc|ccc|c|cc}
      \toprule[1pt]
      \multicolumn{1}{c}{\multirow{1}{*}{}} & 
      \multicolumn{1}{l}{\multirow{2}{*}{Method}} & 
      \multicolumn{7}{c}{Intra-instance} &
      \multicolumn{2}{c}{Inter-instance} &
      \multicolumn{3}{c}{Background} &
      \multicolumn{1}{c}{Others} &
      \multicolumn{2}{c}{Overall} \\
      
      \cmidrule{3-15}\cmidrule{16-17} 
      \multicolumn{1}{c}{\multirow{1}{*}{}} & &
      {\normalsize Blurred} & 
      {\normalsize Artistic} & 
      {\normalsize Glass} & 
      {\normalsize\begin{tabular}[c]{@{}c@{}}Single-\\ Char\end{tabular}} & 
      {\normalsize Inverse} & 
      {\normalsize Distorted} & 
      {\normalsize Delimited} & 
      {\normalsize Dense} & {\normalsize Overlapped} & {\normalsize Occluded} &
      {\normalsize \begin{tabular}[c]{@{}c@{}}Low-\\ Contrast\end{tabular}} & 
      {\normalsize \begin{tabular}[c]{@{}c@{}}Complex-\\ BG\end{tabular}} & 
      {\normalsize Others} &
      {\normalsize Hard} &
      {\normalsize Norm}
      \\
      
      \midrule[1pt]
        \multirow{5}{*}{\rotatebox[origin=c]{90}{Pretrain\hspace{-8pt}}}
           & DBNet++ &
           5.6 & 21.9 & 19.9 & 2.9 & 35.0 & 19.4 & 25.0 & 8.10 & 21.8 & 20.3 & 28.0 & 9.20 & 5.50 & 20.5 & 31.3
          \\
           & ABCNet v2 
           & 13.6 & 21.0 & 19.1 & 7.8 & 23.1 & 9.90 & 13.8 & 17.9 & 15.1 & 26.6 & 45.4 & 14.2 & 7.90 & 27.1 & 46.7
          \\
           & DPText 
           & 13.6 & 23.7 & 23.0 & 8.0 & 23.3 & 12.1 & 16.2 & 18.3 & 14.5 & 26.1 & 45.2 & 16.4 & 10.0 & 27.9 & 48.7
          \\
           & SRFormer &   
             {22.9} & 28.2 & 24.6 & 15.3 & 23.7 & 18.2 & 16.0 & 25.6 & 19.7 & 33.5 & 46.8 & 21.9 & 7.80 & 33.2 & 53.3
          \\

      \midrule 
      \multirow{6}{*}{\rotatebox[origin=c]{90}{Finetune\hspace{-8pt}}} 
      & DBNet & 27.4 & 26.5 & 27.1 & 	2.71 & 43.9 & 27.5 & 13.6 & 36.2 & 23.7 & 31.0 & 50.2 & 11.1 & 8.25 & 34.8 & 59.4 \\
      & DRRG & 7.57 & 17.8 & 30.1 & 4.81 & 58.1 & 25.6 & 40.5 & 5.04 & 21.4 & 27.6 & 34.6 & 21.4 & 10.2 & 24.9 & 35.1 \\
      & FCENet & 22.4 & {41.3} & \textbf{49.2} & 11.9 & \underline{75.6} & \textbf{42.7} & 22.8 & \underline{29.5} & \underline{34.3} & 32.3 & {60.7} & 36.0 & \underline{25.2} & 41.1 & 65.0 \\
      & Mask-RCNN & 24.5 & 17.9 & 18.5 & 6.15 & 30.4 & 13.0 & 19.4 & 27.6 & 26.4 & 32.6 & 42.7 & 7.46 & 9.04 & 31.4 & 53.6 \\
      & PSENet & 17.7 & 15.3 & 15.3 & 3.24 & 16.7 & 14.9 & 12.7 & 27.1 & 21.9 & 25.7 & 35.3 & 1.94 & 3.74 & 25.0 & 48.5 \\
      & TCM & \textbf{35.0} & 35.5 & 7.20 & \underline{42.1} & 37.2 & 19.5 & \textbf{62.5} & \textbf{39.9} & \textbf{36.9} & 38.5 & 58.7 & 23.1 & 15.4 & \underline{43.4} & \textbf{68.4} \\

        \midrule
        \multirow{4}{*}{\rotatebox[origin=c]{90}{Joint\hspace{-10pt}}}
        & DBNet++ 
        & 21.2 & \textbf{44.2} & {49.0} & 17.1 & \textbf{76.1} & \underline{41.8} & 40.7 & 21.6 & {30.6} & 38.3 & \textbf{64.3} & \textbf{42.2} & 22.2 & {45.0} & {66.9} \\
        & PANet &   
             \underline{23.6} & 34.2 & 37.3 & 19.2 & 60.5 & 33.9 & 30.7 & 23.9 & 20.5 & \underline{38.9} & 61.7 & 23.2 & 5.50 & 42.1 & 65.1 \\
        & DPText &
        20.7 & 25.1 & 21.1 & 19.9 & 27.3 & 12.4 & 19.7 & 24.7 & 14.3 & 32.3 & 48.0 & 15.8 & 10.1 & 33.2 & 54.2 \\
        
        \midrule
        \multirow{2}{*}{\rotatebox[origin=c]{90}{SSL\hspace{0pt}}}
        & MAEDet$^\dagger$  & 
            19.6 & 41.1 & 38.4 & 17.0 & 66.0 & 37.4 & 44.2 & 21.7 & 20.4 & 38.1 & 58.6 & \underline{23.4} &  \textbf{31.1} & 41.2 & 63.5 \\
        & MAEDet &
             23.3 &  \underline{43.5} & \underline{45.2} &  \underline{25.1} & 73.3 & 41.1 &  \underline{45.7} &  25.7 &  {28.6} &  \textbf{43.9} & \underline{63.3} & 23.3 & {23.9} &  \textbf{45.7} &  \underline{66.9} \\ 
      \bottomrule[1pt]
    \end{tabular}
  }
  \caption{
  Comparison of MAEDet with other STD models on LTB under different training strategies. F-measure (\%) is reported.
  ``$\dagger$'' denotes model is trained without self-supervised weights. ``SSL'' for self-supervised learning. \textbf{Bold} is 
  the best and \underline{underline} is second best.
}
  \label{tab:eval_on_ltb}
\end{table*}

\subsection{Baseline}
Inspired by CLIP \cite{radford2021learning} and MAE \cite{he2022masked}, we adopt self-supervised learning to address the long-tailed problems from the data perspective, termed as MAEDet. As shown in the \Cref{fig:baseline}, during the stage of SSL, given an image $I \in \mathbb{R}^{h \times w \times 3}$ and its randomly masked version $I_{\text{mask}}$, the model reconstructs the input as $I^{\text{rec}}$. To focus on the text region and model the text feature better, a guidance mask $M_g$ is generated by the frozen CLIP. Specifically, $M_g$ is the attention map, and $M$ is the binary map with $\mathbf{T}$, which is the threshold for discriminating between the text region and background, and ``Text" is the prompt of the text encoder of CLIP. Considering reconstruction bias between the text region and the background, we add an auxiliary branch to optimize MAEDet with the pixel-level balanced reconstruction loss $\mathcal{L}_{br}$,  formulated as \Cref{eq:loss_br}, where $\alpha$ is the balance factor, $\mathds{1}$ is the indicator function, and $\mathcal{L}_2$ is the L2 loss function.
 At the supervised learning stage, MAEDet is initialized from the pretrained backbone and optimized using detection loss $\mathcal{L}_{det}$.

\begin{equation}
\begin{split}
\mathcal{L}_{br}=\sum_{i=1}^H\sum_{j=1}^W 
  &\alpha \mathds{1}_{M_g > \mathbf{T}}  \mathcal{L}_2\left(I_{ij} - I^{\text{rec}}_{ij}\right) \\
  +&(1 - \alpha)  \mathds{1}_{M_g \le \mathbf{T}} \mathcal{L}_2\left(I_{ij} - I^{\text{rec}}_{ij}\right)  
    \label{eq:loss_br}.
\end{split}
\end{equation}

\section{Experiments}

In this section, we will first verify the performances of the representative detectors on mainstream benchmarks with the proposed JDL paradigm. Subsequently, the LTB benchmark is leveraged to evaluate further the long-tailed ability of these models, along with our baseline model MAEDet. 

\subsection{Experimental Setup}
\subsubsection{Implement Details}
We use three different training strategies: 1) Pretrain: We utilize officially pretrained models for each detector, typically initialized with ImageNet pretraining weights; 2) Joint: Models are trained from scratch on the union of previous English word-level training sets, termed Joint98K, without loading any pretrained weights; 3) SSL: Our proposed MAEDet incorporate self-supervised learning using MARIO-LAION \cite{chen2024textdiffuser} before training on Joint98K.

\begin{table}[!tb]
\tiny
\centering
\renewcommand{\arraystretch}{1.22}
\resizebox{\linewidth}{!}{
\begin{tabular}{clccc}
    \toprule
    \multicolumn{1}{c}{\multirow{2}{*}{\# Exp.}} & \multicolumn{1}{l}{\multirow{2}{*}{Setup}} & \multicolumn{2}{c}{LTB} & Joint98K \\
    \cmidrule(lr{3pt}){3-4}\cmidrule(lr{3pt}){5-5}
    & & Hard & Norm & Avg. \\

    \midrule
    \multicolumn{5}{l}{\textit{Input Size}} \\
    
    1 & 480 & 34.3 & 53.5 & 63.8 \\
    2 & 640 & 41.0 & 62.9 & 71.3 \\
    3 & 800 & \textbf{43.8} & \textbf{67.5} & \textbf{74.4} \\
    
    \multicolumn{5}{l}{\textit{Mask Threshold}} \\
    
    4 & $\mathbf{T}=0.1$ & \textbf{44.8} & 66.3 & \textbf{74.2} \\
    5 & $\mathbf{T}=0.2$ & 44.7 & \textbf{66.4} & 73.9 \\
    6 & $\mathbf{T}=0.3$ & 44.2 & 65.8 & 73.7 \\

    \multicolumn{5}{l}{\textit{Optimization objective}} \\

    7 & $\mathcal{L}_{rec}$ & 45.4 & 66.4 & 74.1 \\
    8 & $\mathcal{L}_{rec} + \mathcal{L}_{br}$ & \textbf{45.7} & \textbf{66.9} & \textbf{74.5} \\

    \multicolumn{5}{l}{\textit{Balance Factor}} \\

    9 & $\alpha=0.8$ & 42.0 & 63.5 & 71.6 \\
    10 & $\alpha=0.9$ & \textbf{45.7} & \textbf{66.9} & \textbf{74.5} \\
    11 & $\alpha=1.0$ & 44.7 & 66.4 & 73.9 \\
    \bottomrule
\end{tabular}
}
\caption{ 
    Ablation study results for MAEDet using Joint-Dataset Learning. 
}
\label{tab:ablation}
\end{table}
\subsubsection{Evaluation Protocols}
We employ the F-measure \cite{karatzas2015icdar} as the standard evaluation metric, focusing solely on text instances labeled as ``care'' and disregarding ``don't care'' regions. For the LTB dataset, we report the F-measure for each of its 13 categories. Additionally, we use two overall metrics: \textit{Hard}, which marks only challenging text instances as ``care,'' and \textit{Norm}, which retains the original ground truth annotations. These metrics provide complementary insights, with the Hard metric better differentiating model performance in challenging scenarios.

\subsection{Main Results and Analysis}

\Cref{tab:eval_on_9_datasets} presents a comparison between them. 
DBNet++ and DPText-DETR, when evaluated with the proposed JDL paradigm, achieve higher performance on more challenging datasets (COCO, ArT, LSVT, and TextOCR) than ``Pretrain" settings. Additionally, the pretrained SRFormer exhibits excessively high performance on Total-Text, MLT2017, and MLT2019, likely due to data leakage, where its pertaining data includes some validation images from Joint98K.

Moreover, we also verify the effectiveness of JDL in reducing over-fitting in STD evaluation protocol between. As observed in \Cref{tab:finetuning_gap}, when DPText-DETR is fine-tuned on Total-Text, a notable performance gap of 15.1\% emerges between its performance on ICDAR2015 and Total-Text.
However, this discrepancy is significantly reduced to merely 6.0\% when employing the JDL protocol. This substantial reduction in performance variance across different datasets substantiates the effectiveness of JDL in enhancing model generalization and alleviating the dataset-specific overfitting issue. 

\Cref{tab:eval_on_ltb} compares the different strategies on LTB. We observe that models trained with Joint98K outperform those pretrained ones, demonstrating that joint-dataset training with extensive real data better addresses long-tailed distribution challenges by more accurately reflecting real-world conditions.

\subsection{Ablation Studies} 

Exp. 1-3 in \Cref{tab:ablation} demonstrate that appropriately increasing the input size for the LTB is beneficial for capturing challenging text instances. 
Mask threshold $\mathbf{T}$ in \Cref{eq:loss_br} determines the proportion of text the backbone perceives during the self-supervised pretraining stage.  Exp. 4-6 \Cref{tab:ablation} show the best performance occurs when $\mathbf{T} = 0.1$. We hypothesize that a lower $\mathbf{T}$ allows more text pixels to contribute to the loss calculation. 
Exp. 7-8 shows that after optimizing pretraining with $\mathcal{L}_{br}$, there is a significant improvement in performance from LTB to Joint98K, demonstrating that our auxiliary loss is effective in representing text features for scene text detection.
The balance factor $\alpha$ controls the proportion of text and non-text regions perceived during the self-supervised learning stage. Exp. 9-11 demonstrate that MAEDet achieves the best performance when $\alpha=0.9$, suggesting that some background pixels should be included in the comprehensive optimization objective. Therefore, we use $\alpha=0.9$ as the default settings.

\section{Conclusion}
The paper identifies the fine-tuning and long-tailed problems as two significant factors contributing to the discrepancy between scene text detector performance on academic benchmarks and practical applications. 
Experiments demonstrate that models optimized for a single dataset often overfit specific domains but struggle to generalize to real-world scenarios. Given that Dataset-Specific Optimization (DSO) leads to such fine-tuning gap, we advocate Joint-Dataset Learning (JDL) for practical demands.
Regarding the long-tailed problems, we define 13 categories of long-tailed challenges and introduce the Long-Tailed Benchmark (LTB), the first holistic benchmark to evaluate model's ability to detect different types of challenging texts, setting a new standard in scene text detection. Additionally, MAEDet is proposed as the baseline for our benchmark.
We hope this work could provide valuable insights into improving scene text detectors and envision more efforts being devoted to practical approaches that effectively address long-tailed problems.

\section*{Acknowledgments}
Supported by the National Natural Science Foundation of China (Grant NO 62376266 and 62406318), Key Laboratory of Ethnic Language Intelligent Analysis and Security Governance of MOE, Minzu University of China, Beijing, China.

\newpage
\bibliographystyle{named}
\bibliography{ijcai25}

\clearpage
\appendix

\section{Powerlessness of Scene Text Detector}
\begin{figure}[!h]
\centering
\includegraphics[width=0.48\textwidth]{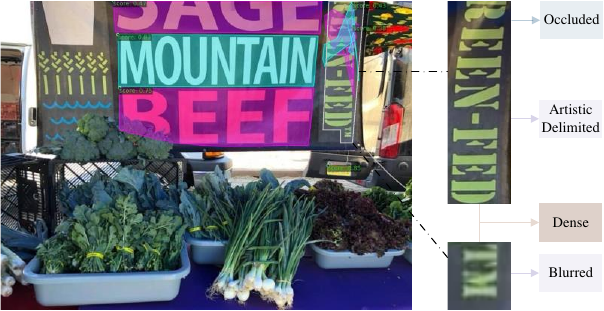}
    \caption{Detection results of DPText-DETR in practical applications. 
    Left: a complete image with the visualization of detection masks. 
    Middle: the text instances that are not detected, shown in an enlarged view for better visibility.
    Right: challenge attributes of text instances. 
    Purple, brown, and blue indicate intra-instance problems, inter-instance problems, and background problems, respectively.}
    \label{fig:applications}
\end{figure}

Real-world scene text is inherently complex, as a single text instance may possess multiple challenging attributes that stump detectors. As illustrated in \Cref{fig:applications}, the text ``reen-fed'' simultaneously belongs to the \textit{occluded}, \textit{artistic}, and \textit{delimited} categories, while ``TM'' is \textit{blurred} due to limited image quality. Both of them belong to the \textit{dense} category for being closely positioned. Similar examples abound, indicating the powerlessness of current detectors in handling such multifaceted challenges.

\section{More analysis of Training and Evaluation Protocol}
\label{appd:protocol}

\subsubsection{The Imperative of Joint-Dataset Learning}

\begin{figure}[!b]
\centering
\includegraphics[width=0.48\textwidth]{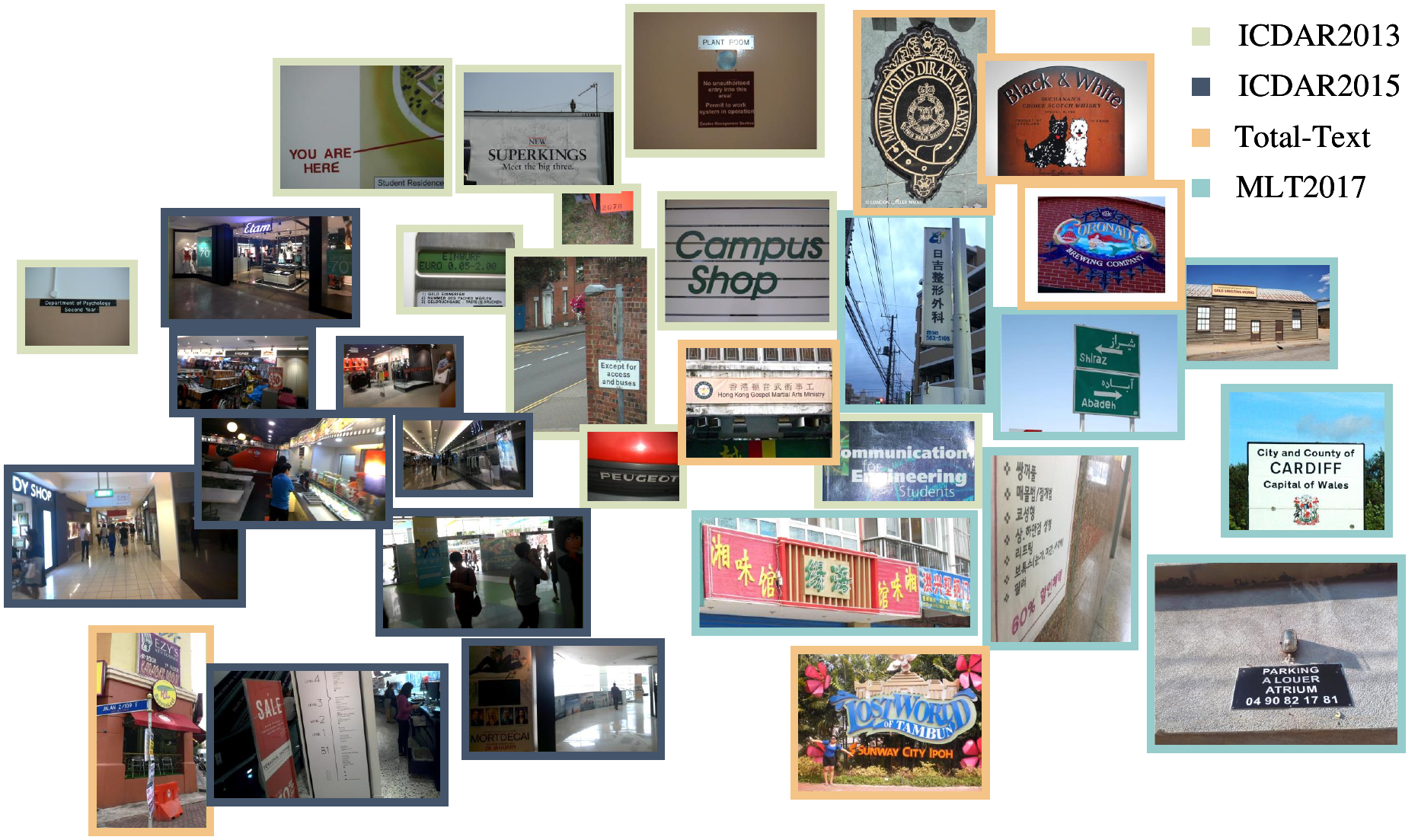}
    \caption{
    A visualization of clustering results for images derived from four distinct datasets. Features are extracted using the oClip method \protect\cite{xue2022language}, and t-SNE is employed to map the feature distribution, with individual clusters accentuating the inherent differences among the datasets.
    }
    \label{fig:dataset}
\end{figure}

In the domain of scene text detection, datasets exhibit significant domain differences, arising from variations in the real-world contexts they represent, as well as differences in text features such as font, layout, and background complexity, which stem from distinct data collection practices. These disparities are further detailed in \Cref{appd:dataset_details}. Such variations introduce significant dataset biases, making it difficult for models to generalize effectively across diverse scenarios. Models trained on a single dataset often encounter performance degradation when evaluated on others. To illustrate this, we visualize the clustering results of images from four distinct datasets in \Cref{fig:dataset}, which demonstrate that even basic clustering methods can easily differentiate between these datasets. For example, ICDAR2015 primarily contains street-view images captured in commercial areas, often with blurry text and varying orientations. In contrast, TotalText consists largely of trademark and emblem images, where text is frequently presented in curved or highly stylized forms. These domain shifts pose significant challenges when employing the traditional DSO approach. Models trained on a single dataset tend to overfit to the unique characteristics of the training data, limiting their ability to generalize across different scenarios. Consequently, this overfitting results in performance degradation when the model is applied to datasets with distinct characteristics from the training set.

Joint-Dataset Learning offers a promising solution to address these limitations. By training detectors on the combined diversity of multiple datasets, JDL enables the detectors to learn more robust and generalized features that capture a wider range of text types, layouts, and environmental factors. JDL not only mitigates overfitting but also enhances the model's adaptability to the complexities of real-world scenarios, where text can appear in various forms and contexts. Consequently, JDL provides a more effective and scalable framework for scene text detection, addressing the key challenges posed by domain shifts and dataset-specific biases.

Furthermore, models trained using JDL are systematically evaluated across multiple benchmarks in a unified manner. This comprehensive evaluation methodology ensures a more objective and fair assessment of the model's generalization capability, facilitating a deeper understanding of its performance across diverse conditions.

\subsubsection{The Feasibility of Joint-Dataset Learning}

Scene text detection is a precise task that focuses on localizing text within an image by predicting the coordinates of the text regions. A key characteristic of STD is that the data format is typically standardized. Specifically, each text instance is represented as a set of coordinates that precisely define the location of the text within the image. This structured format enables a consistent representation across different datasets, making it feasible to combine data from multiple sources for joint training without introducing significant complexity.

Unlike tasks with highly variable data structures, such as object detection, where the data may involve diverse object categories with different feature representations, STD datasets share a common organizational structure. This uniformity allows for the straightforward merging of datasets that may vary in factors such as text orientation, font, and layout. By aggregating these datasets, the model is exposed to a broader range of real-world variations, facilitating the learning of more generalized features. Consequently, training on such a combined dataset enhances the model's ability to generalize across different text forms, backgrounds, and environmental conditions, thereby improving its robustness and performance on unseen data distributions.

While scene text detection applies to various scenarios, its core objective—detecting and localizing text within natural scenes—remains consistent. This uniformity facilitates the merging of data from diverse datasets without significant complexity, as the task focuses on accurately identifying text regions regardless of variations in text types. In contrast, tasks in other domains often experience more substantial domain differences, which hinder the direct merging of datasets. For instance, in tasks like Machine Translation, the discrepancy between source and target languages in terms of syntax, vocabulary, and grammar creates challenges when combining datasets from different languages. Similarly, in specialized domains such as medical or legal translation, the need for domain-specific knowledge and terminology further complicates dataset integration. These domain-specific variations make JDL less feasible, as the models must adapt to drastically different feature representations and contextual requirements.

\subsubsection{Similar Protocols in Other Deep Learning Domains}

Joint-Dataset Training is not exclusive to the domain of scene text detection. Similar strategies have been adopted across various fields, where the conventional practice of single-dataset optimization has been supplanted by more generalized approaches that enhance model robustness and generalization. Below are several pertinent examples:

\begin{itemize}
    \item \textbf{Advancements in Scene Text Detection:} Recently, several state-of-the-art scene text detection methods have moved beyond the traditional DSO framework, incorporating a joint-dataset training strategy akin to JDL. Notably, the Masked Text Detection framework proposed by Wang et al. and the Masked Text Spotting method by He et al. both utilize joint training on multiple datasets. This approach mitigates the risk of overfitting to a single dataset, leading to substantial improvements in model performance across diverse text instances and settings.

    \item \textbf{Pretraining in Image-Text Retrieval:} In the domain of image-text retrieval, models are tasked with understanding the interplay between visual and textual modalities. Approaches such as CLIP (Contrastive Language-Image Pretraining) \cite{radford2021learning} and its derivative models, have been pretrained on extensive collections of image-text pairs, thereby obviating the need for separate fine-tuning for individual retrieval tasks. This pretraining paradigm significantly boosts the model's ability to generalize, as evidenced by its evaluation across a variety of retrieval benchmarks, showcasing its adaptability to diverse datasets.

    \item \textbf{Large Language Models:} Models like GPT, which are trained on vast, general-purpose datasets without task-specific fine-tuning, exemplify the benefits of generalized training approaches. These models are assessed on multiple benchmarks for general-purpose question-answering and text-generation tasks, underscoring their capacity for generalization and efficient adaptation to a range of downstream applications.
\end{itemize}

The success of JDL and similar strategies across various domains demonstrates their effectiveness in enhancing model performance while addressing issues such as overfitting and data bias. In the context of STD, the adoption of JDL facilitates evaluation across a wide spectrum of text scenarios and diverse datasets, thereby enabling a more robust assessment of a model's generalization ability and its applicability in real-world conditions.

\section{Dataset Details}
All models are evaluated on our proposed Joint98K and LTB benchmarks: the former demonstrates the existence of the fine-tuning gap and the latter tests the ability of models to detect long-tailed text instances.

\subsection{Joint98K}\label{appd:dataset_details}
We use Joint98K for joint training and evaluation with the JDL paradigm.
Out of considerations for English text instances annotated at the word level, Joint98K combines the following nine datasets:

\begin{table}[!b]
\centering
\small
\begin{tabularx}{\linewidth}{l*{5}L}
      \toprule[1px]
      \multirow{2}{*}{Datasets} & 
      \multicolumn{3}{c}{\# Original} &
      \multicolumn{2}{c}{\# Ours} \\
      \cmidrule{2-6}
      \multicolumn{1}{c}{\multirow{1}{*}{}} & \multicolumn{1}{c}{\multirow{1}{*}{Train}} & \multicolumn{1}{c}{\multirow{1}{*}{Val}} & \multicolumn{1}{c}{\multirow{1}{*}{Test}} & \multicolumn{1}{c}{\multirow{1}{*}{Train}} & \multicolumn{1}{c}{\multirow{1}{*}{Test}} \\
      
      \midrule[1px]
      ICDAR2013$^*$ & 229 & \begin{tabular}[c]{c}-\end{tabular} & 233 & 229 & 233 \\
      ICDAR2015$^*$ & 1000 & \begin{tabular}[c]{c}-\end{tabular} & 500 & 1000 & 500 \\
      COCO-Text$^*$ & 43686 & 10000 & 10000 & 43686 & 10000 \\
      Total-Text$^*$ & 1255 & \begin{tabular}[c]{c}-\end{tabular} & 300 & 1255 & 300 \\
      MLT2017$^{\dagger\ddagger}$ & 9000 & \begin{tabular}[c]{c}-\end{tabular} & 9000 & 785 & 197 \\
      MLT2019$^{\dagger\ddagger}$  & 10000 & \begin{tabular}[c]{c}-\end{tabular} & 10000 & 800 & 200 \\
      ArT$^{\dagger}$ & 5603 & \begin{tabular}[c]{c}-\end{tabular} & 4563 & 4482 & 1121 \\
      LSVT$^{\dagger}$ & 30000 & - & 2000 & 24000 & 6000 \\
      TextOCR$^*$ & 21778 & 3124 & 3232 & 21778 & 3124 \\
      
      \midrule
      Total & - & - & - & 98015$\downarrow$ & 21675 \\
      
      \bottomrule[1px]
\end{tabularx}

    \caption{
    Composition and distribution of Joint98K. $*$ denotes the original training and validation sets that are directly adopted as our training and test sets without any modifications. $\dagger$ denotes the original training set is split into our training and test sets with an 8:2 ratio due to missing test annotations. $\ddagger$ denotes that only images containing English text are selected. $\downarrow$ denotes only images with at least one text instance are used, reducing the total count of training images.}
\label{tab:joint_training_data}
\end{table}

\begin{itemize}[leftmargin=1em]
    \item The Focused Scene Text of \textbf{ICDAR2013} is designed to advance scene text detection and recognition. The images primarily depict horizontal English text in indoor and outdoor settings,  with ``focused'' indicating texts explicitly focused by photographers. 
    
    \item The Incidental Scene Text of \textbf{ICDAR2015} differs from the focused one in that its images are collected freely without any actions to improve the position or quality of texts in the frame deliberately. It includes street scene images featuring incidental texts, which vary in orientation.

    \item \textbf{COCO-Text} is a large-scale dataset aiming to enhance the complexity of text detection tasks by introducing a broader and more challenging range of scenarios. Text instances are meticulously annotated using polygons with multiple points rather than quadrilaterals, enabling precise boundary delineation for texts of curved and irregular text orientations. 
    
    \item \textbf{Total-Text} is designed to address challenges associated with curved text. Most detectors designed before 2017 assume that scene text is horizontally aligned before 2017, while real-world text layouts include vertical, curved, and even circular forms. Over half of the texts in Total-Text exhibit curved forms. 

    \item \textbf{MLT2017} \& \textbf{MLT2019} stand as the pioneering dataset designed explicitly to test multi-lingual text detection methodologies and to foster advancements in script-robust text detection methods.

    \item \textbf{Ar}bitrary Shaped \textbf{T}ext (\textbf{ArT}) aims to offer a more extensive benchmark of curved and irregularly shaped text in complex scenes. Curved and irregularly shaped text is prevalent in real-world scenarios. However, Total-Text and SCUT-CTW1500 provide only 800 test images together, which is inadequate for comprehensively evaluating detectors. 

    \item The \textbf{L}arge-scale \textbf{S}treet \textbf{V}iew \textbf{T}ext with Partial Labeling (\textbf{LSVT}) offers a larger scale of fully annotated images, which are sourced from diverse street scenes in China.

    \item \textbf{TextOCR} focus on high-text-density scene text tasks. Most datasets are relatively small or primarily intended for outdoor or storefront scenes. Moreover, they typically feature fewer words per image, making them less dense and diverse. TextOCR addresses these limitations by providing a large-scale dataset with an average of 32 words per image.
\end{itemize}

Joint98K consists of 98,015 training images, each containing at least one English text instance, and a combined test set of 21,675 images, as detailed in \Cref{tab:joint_training_data}.

\begin{table*}[!t]
  \centering
  \small
  \begin{tabularx}{\linewidth}{@{}
    >{\raggedright\arraybackslash}p{2.1cm}@{}
    C
    C
    C
    C
    >{\hsize=0.8\hsize\centering\arraybackslash}X
    >{\hsize=0.8\hsize\centering\arraybackslash}X
    C
    C
    >{\hsize=0.8\hsize\centering\arraybackslash}X
    >{\hsize=1.1\hsize\centering\arraybackslash}X
    C
    >{\hsize=0.7\hsize\centering\arraybackslash}X
    >{\hsize=0.5\hsize\centering\arraybackslash}X@{}
    >{\hsize=1\hsize\centering\arraybackslash}X
    @{\hspace{0.5cm}}X
  }
       
      \toprule[1pt]
      \multirow{3}{*}{Datasets} & 
      \multicolumn{13}{c}{\# Image Numbers / \# Instance Numbers} \\
      \cmidrule(l){2-15} &
      \multicolumn{7}{c}{Intra-instance} &
      \multicolumn{2}{c}{Inter-instance} &
      \multicolumn{3}{c}{Background} &
      \multicolumn{1}{c}{Others} &
      \multicolumn{1}{c}{\multirow{2}{*}{Total}} \\
      \cmidrule(l){2-14} &
      1 & 
      2 & 
      3 & 
      4 & 
      5 & 
      6 & 
      7 & 
      \multicolumn{1}{|c}{8} & 9 & 
      \multicolumn{1}{|c}{10} & 
      11 & 
      12 &
      \multicolumn{1}{|c}{13} &
      \\

      \midrule[1pt]
      NightTime-ArT &   
        -& 
        -&
        -&
        -&
        -&
        -&
        -& 
        - &
        -& 
        - & 
        46/390 &   
        - &  
        - &  
        46/390
      \\
      Total-Text &   
        40/66 &  
        12/13 &
        18/21 &
        5/5 &
        2/5 &
        2/2 &
        - &
        32/40 &
        1/1 & 
        18/21 &
        7/10 &
        3/3 &  
        7/7 &  
        72/114 
        \\
      ICDAR2015 &   
        134/200 & 
        28/31 &
        1/1 &
        39/54 &
        18/24 &
        1/1 &
        -& 
        56/91 &
        2/3 & 
        35/39 & 
        29/38 &
        16/19 &   
        5/5 &  
        168/264 
      \\
      ORT &   
        89/169 & 
        46/95 &
        7/8 &
        1/1 &
        6/7 &
        29/47 &
        - & 
        29/73 &
        1/2 & 
        201/787 & 
        16/26 &   
        4/4 &  
        1/2 &  
        201/815 
      \\
      Inverse-Text &   
        44/132 & 
        57/153 &
        38/47 &
        30/103 &
        23/37 &
        7/9 &
        43/120 & 
        13/87 &
        10/25 & 
        64/122 & 
        32/117&   
        17/44 &  
        2/2 &  
        227/628
      \\
      ArT &   
        147/395 & 
        55/119 &
        35/56 &
        4/5 &
        17/26 &
        8/8 &
        - & 
        103/240 &
        5/9 & 
        29/43 & 
        27/43 &   
        11/14 &  
        27/35 &  
        210/559
      \\
      \midrule
      Total &   
        454/962 & 
        198/411 &
        99/133 &
        79/168 &
        66/99 &
        47/67 &
        43/120 & 
        225/522 &
        19/40 & 
        347/1012& 
        157/624 &   
        51/84 &  
        42/51 &  
        924/2770 
      \\
      \bottomrule
  \end{tabularx}
  \caption{Distribution of challenging text instances across different datasets of LTB, categorized by predefined standards. Columns 1-13 correspond to 13 long-tailed challenges, which are blurred, artistic, glass, single-character, distorted, inverse, delimited, dense, overlapped, occluded, low-contrast, complex-background, and other. \textit{\# Image Numbers} denotes the number of images that contain at least one instance of a particular challenge category. \textit{\# Instance Numbers} denotes the total number of text instances that belong to a specific challenge category.}
  \label{tab:benchmark_distribution}
\end{table*}

\subsection{LTB}

\subsubsection{Data Collection}
\begin{figure}[!t]
  \centering
  \includegraphics[width=0.48\textwidth]{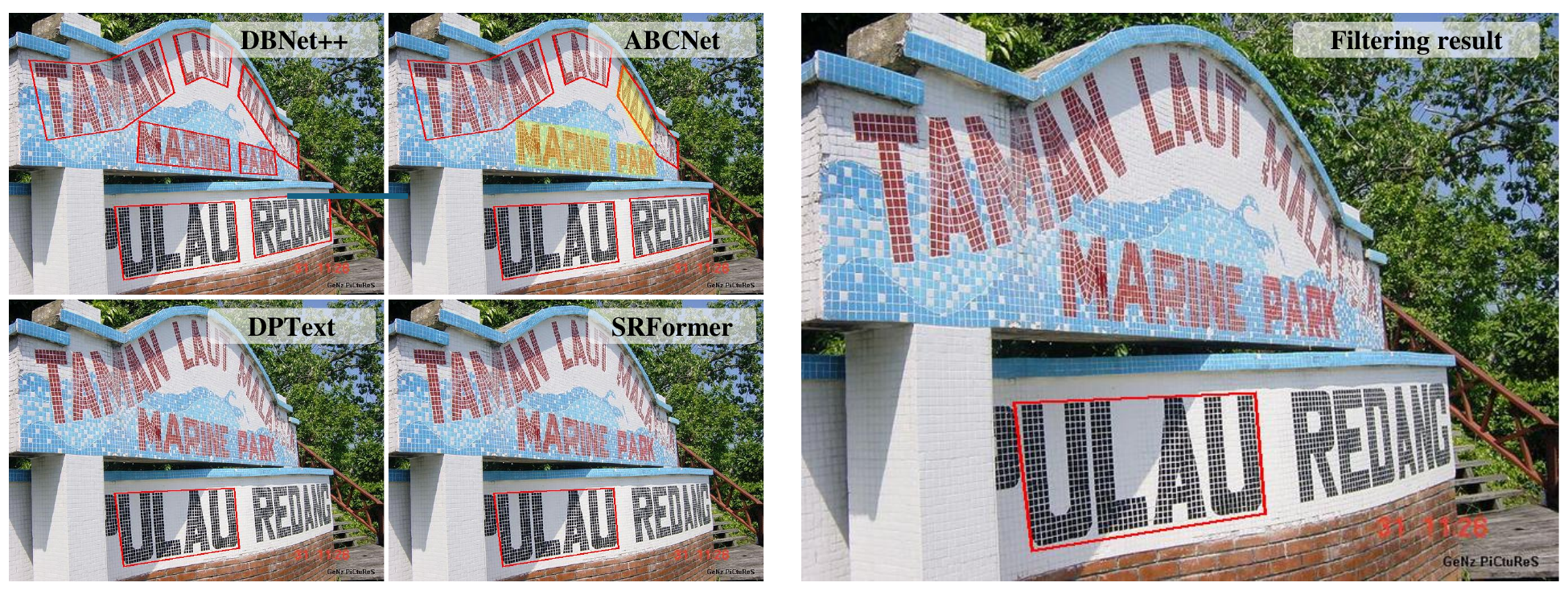}
  \caption{A visualization of the filtering process. 
  Red bounding boxes mark the undetected text instances. Yellow masks denote the detected regions that have a small intersection with the undetected instances so that these detection results are considered failed.
  }
  \label{fig:filtering_process}
\end{figure}

Based on the criteria and strategy described in \Cref{sec:ltb_pipline}, LTB is composed of data from six datasets. The data from ICDAR2015, Total-Text, and ArT are selected with the assistance of detectors. To screen for challenging text instances, we filter out all instances detected by any detectors, as illustrated in \Cref{fig:filtering_process}. Detailed information about these datasets can be found in \Cref{appd:dataset_details}. The rest of the data are manually collected from another three datasets. Below, we outline the characteristics of them:
\begin{itemize}[leftmargin=1em]

    \item \textbf{Inverse-Text} includes numerous instances of inverse text. This dataset is particularly challenging due to the mirror-reversed characters that require robust detection algorithms.

    \item \textbf{Occluded RoadText} includes a large number of street scenes with occluded text.
    
    \item \textbf{NightTime-ArT} is designed to evaluate the generalization of the detector on out-of-distribution data. The images are processed from the ArT dataset to represent scenes under low-light nighttime conditions.
\end{itemize}
We then develop an algorithm that compares the model's predictions against the ground truth annotations to filter out undetected text instances. As described in \Cref{alg:filter_undetected_text}, it firstly uses all detectors to infer each image to get detection results. These prediction results are then aggregated and compared against the ground truth annotations. Specifically, the algorithm calculates the Intersection over Union (IoU) for each ground truth text instance with each predicted bounding box. If the maximum IoU for a given \( gt \) is below a predefined threshold, it is considered undetected by all detectors.
\begin{algorithm}[!t]
  \small
  \begin{algorithmic}[1]
    \Require 
    \Statex Detectors: $D$, Images: $I$, Ground Truth: $G$, IoU threshold $t$
    \Ensure 
    \Statex Images with Undetected Instances: $I’$
    \Statex Undetected Instances: $T$
    \Function{Filter}{$D$, $I$, $G$}
      \State $I' \gets \emptyset$; $T \gets \emptyset$
      \For{$i \in I$}
        \State $J \gets$ \textbf{JointPredict}$(D, i)$ 
        \State \Comment{\parbox[t]{.8\linewidth}{Combine prediction results from multiple detectors}}
        \State $G' \gets$ \textbf{GetCareInstance}$(G, i)$ 
        \State \Comment{\parbox[t]{.8\linewidth}{Extract annotations for text regions marked as \textit{care} or \textit{legible}}}
        \State $T' \gets \emptyset$ 
         
        \For{$g \in G'$}
          \State $u \gets$ \textbf{CalcMaxIoU}$(J, g)$     
          \If{$u < t$}
            \State $T' \gets T'\cup g$
          \EndIf
        \EndFor
          
        \If{$T' \neq \emptyset$}
          \State $I' \gets I' \cup i$; $T \gets T'$
        \EndIf
    \EndFor
    \State \Return{$I'$, $T$}
    \EndFunction
    
  \end{algorithmic}
  \caption{Filter Undetected Instances}
  \label{alg:filter_undetected_text}
\end{algorithm}

\subsubsection{Data Processing}
\label{appd:data_processing}
Despite the elaborate collection of data, simply combining them proves inadequate due to inconsistencies in annotation level and the presence of duplicate, non-Latin, and duplicated samples since our research focuses on English text detection at the word level. 
To enhance the quality and relevance of the benchmark, we employ the following strategies:

\textbf{Non-Latin text removal}: We exclude all text instances that contain characters outside the Latin script. For clarity, the Latin script includes standard English letters, numbers, and punctuation marks. 
 
\textbf{Standardization of annotation granularity}: According to the COCO-Text definition, a \textit{word} is an uninterrupted sequence of characters separated by spaces. To ensure consistency with this definition, we undertake two tasks. First, we re-annotate all instances initially labeled as ``occluded English" to the word level for Occluded-RoadText. Second, we manually excluded text instances that are not annotated at the word level. These steps ensure consistent word-level annotations across all datasets for our research.

\textbf{Image deduplication}: To remove duplicate and near-duplicate images across different datasets, we utilize a CNN-based method to eliminate images with a similarity score exceeding 95\%.

As detailed from \Cref{tab:benchmark_distribution}, LTB contains 1,012 occluded text instances (Column 1) and 960 blurred text instances (Column 10), both of which account for more than one-third of the total number of challenging text instances.
According to the original annotations, the number of text instances labeled as \textit{care} and \textit{don't care} is 8,067 and 8,495, which becomes 2,770 and 13,792 in LTB, as LTB only focuses on challenging text instances.
The 2,770 challenging text instances in LTB contain a total of 4,293 challenge attributes, indicating that each text instance involves approximately 1.5 challenge attributes on average. This statistic highlights the complexity of LTB, whose text instances can be subjected to multiple difficult challenges like those in the real world.

\begin{figure*}[t]
  \centering
  \includegraphics[scale=0.85]{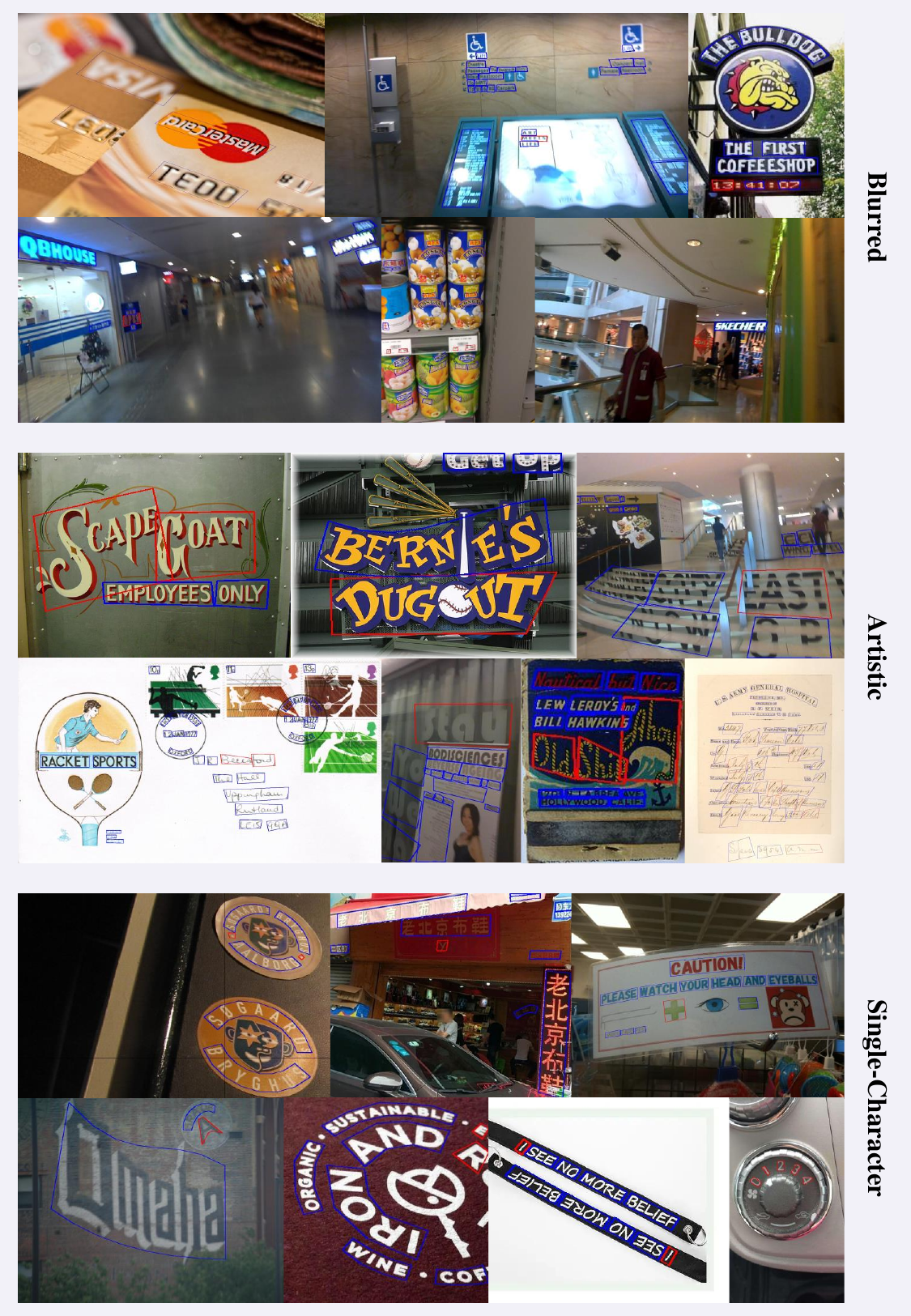}
  \caption{More examples about the inter-instance challenges of LTB. Red bounding boxes mark challenging text instances and blue ones for \textit{don't care} regions.
  }
  \label{fig:more_ltb_examples_1}
\end{figure*}

By incorporating diverse text types and scenarios—ranging from blurred and distorted text to occluded and inverse text, as well as low-light conditions—LTB not only provides a wide variety of test cases but also reduces the bias commonly found in other datasets. With a balanced representation of both common and long-tailed instances, LTB ensures a more accurate evaluation of the ability of detectors in real-world applications, making it a credibility benchmark for evaluating and improving scene text detection techniques. \Cref{fig:more_ltb_examples_1}-\Cref{fig:more_ltb_examples_4} provide more examples about the challenges of LTB at the end of the appendix.

\section{Training Details}
\subsection{Hyper-parameter of MAEDet}
\label{appd:setting}
During the self-supervised learning stage, we follow the default settings of MAE. $\alpha$, and $T$ are set as 0.9, and 0.1 empirically. We pre-train the model on MARIO-LAION for 10 epochs with an AdamW optimizer. The pretraining image size is set to 640 $\times$ 640. The batch size is set to 16 and the learning rate settings and adjustment scheduler are followed MAE \cite{he2022masked}. pretraining is conducted with 8 NVIDIA RTX 2080Ti GPUs. As for the supervised learning stage, we employ an AdamW optimizer with a weight decay of 1e-4. The learning rate is initialized at 0.001 and adjusted by the policy with a power of 0.9. The batch size is set to 16 and the epoch is 100. The joint training is conducted with 4 NVIDIA RTX 4090 GPUs. 


\begin{figure*}[!ht]
\centering
\includegraphics[width=0.98\textwidth]{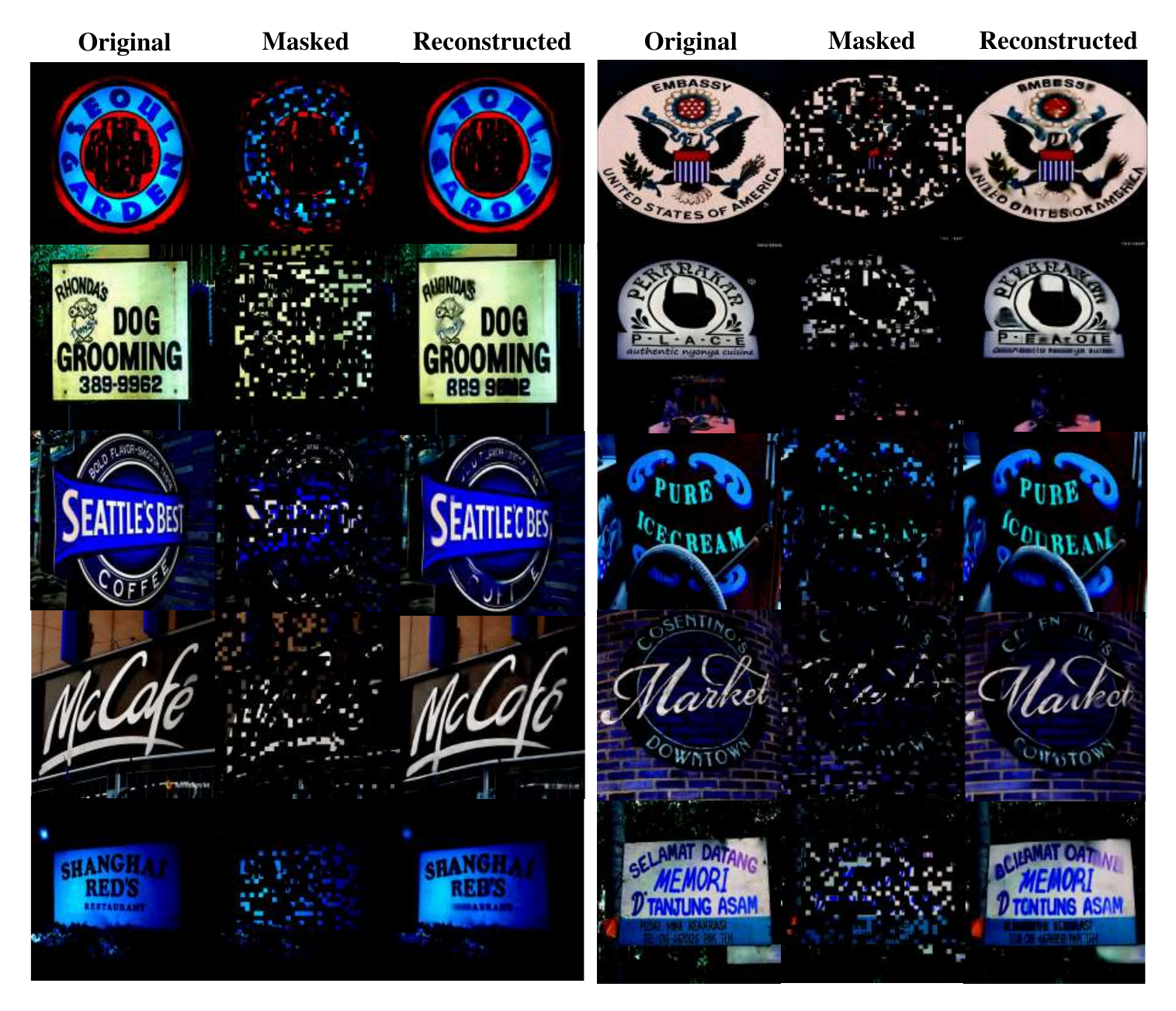}
    \caption{Reconstruction results of MAE.}
    \label{fig:mae}
\end{figure*}

\subsection{Detailed Architecture of MAEDet}
Like other scene text detectors, MAEDet includes tree components, backbone, neck, and detection head. 
\subsubsection{Backbone}
We follow the default settings of ViT-B \cite{2021An}, pretrained with MAE. 
A patch embedding layer and a position embedding layer are applied to prepare the input data for the transformer architecture.
Patch embedding aims to convert the image into the sequence tokens, which adapt the input of ViT. Specifically, given the scene image $I \in \mathbb{R}^{H \times W \times C}$, we first resize it into $640 \times 640$
to balance computational cost and input resolution. We then use a patch embedding layer to split the image into non-overlapping patches, each of size $16 \times 16$. Finally, a linear layer is leveraged to map the patches into a high-dimension space with dimension $d$ of 768.





\subsubsection{Neck}
After extracting the multi-scale features by the backbone, the neck module fuses these features and feeds them into the detection head. Since the vanilla ViT is non-hierarchical, all the feature maps in the backbone are of the same resolution. In this work, We adopt the ``SimpleFPN" in ViTDet to facilitate the multi-scale features in the detection task.  Specifically,
with the default ViT feature map of a scale of $\frac{1}{16}$, we produce feature maps of scale $\{\frac{1}{32}, \frac{1}{16}, \frac{1}{8}, \frac{1}{4}\}$ using convolution and deconvolution layers.

\subsection{Reconstruction Result of MAEDet}
We visualize the reconstruction results during the pretraining stage as shown in \Cref{fig:mae}. These results demonstrate that the backbone has learned rich and discriminative features that enable the model to distinguish between text and non-text regions. However, the reconstruction ability for small texts, potentially limited by the resizing process, still has a large room for improvement.

\begin{figure*}[t]
  \centering
  \includegraphics[scale=0.85]{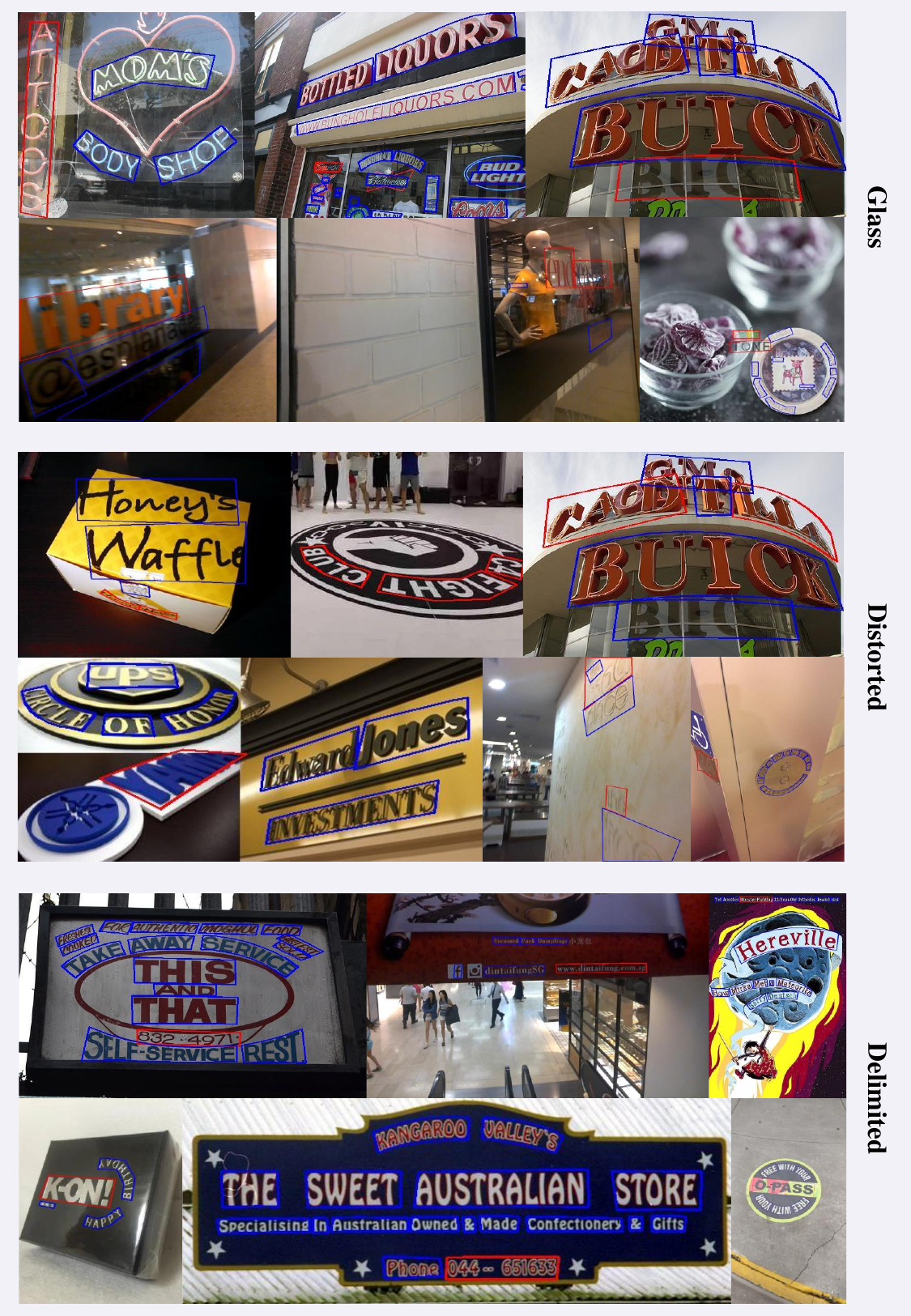}
  \caption{More examples about the inter-instance challenges of LTB. Red bounding boxes mark challenging text instances and blue ones for \textit{don't care} regions.
  }
  \label{fig:more_ltb_examples_2}
\end{figure*}

\begin{figure*}[t]
  \centering
  \includegraphics[scale=0.85]{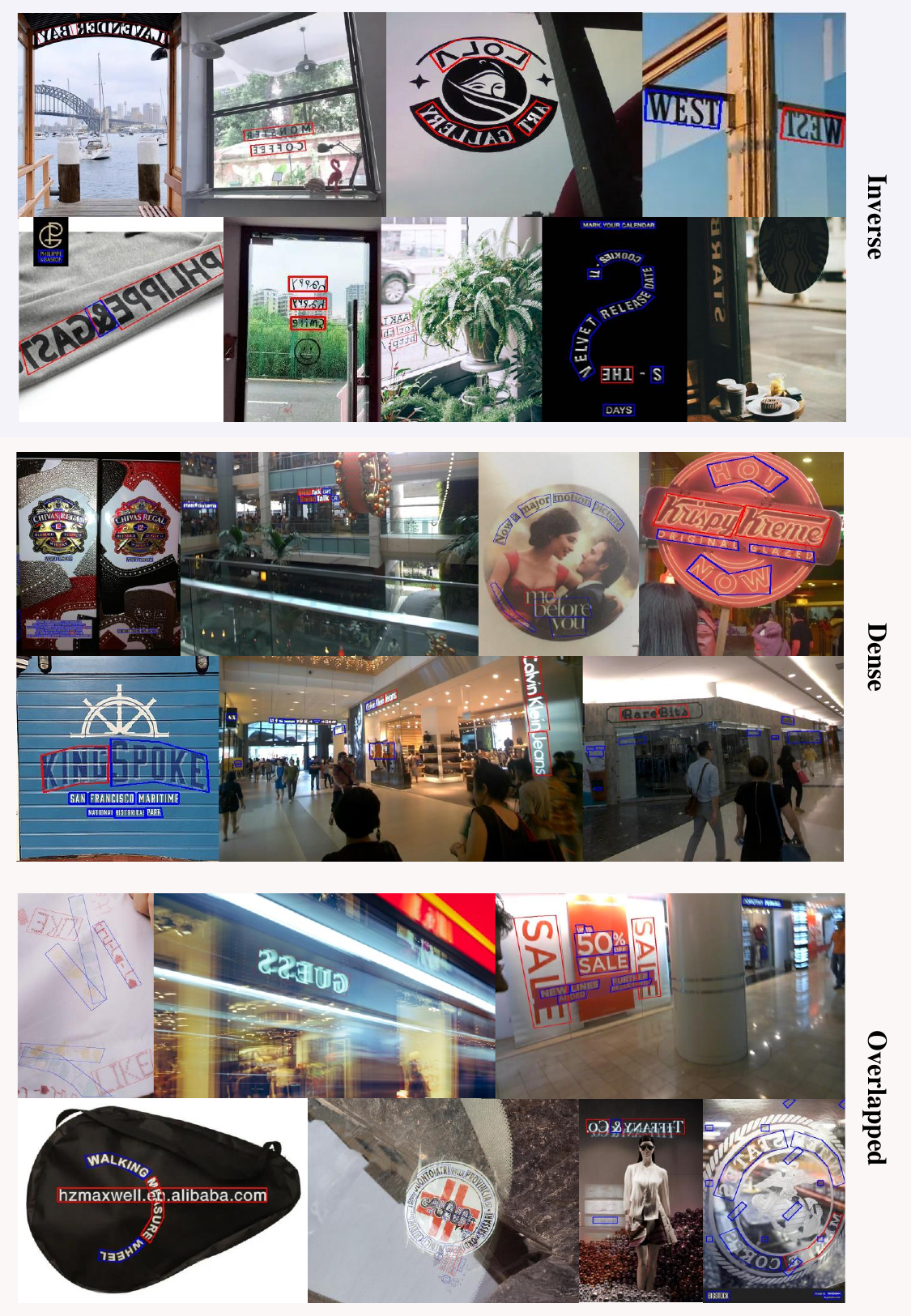}
  \caption{More examples about LTB. Purple and brown denote the inter-instance challenges and intra-instance challenges. Red bounding boxes mark challenging text instances and blue ones for \textit{don't care} regions.
  }
  \label{fig:more_ltb_examples_3}
\end{figure*}

\begin{figure*}[t]
  \centering
  \includegraphics[scale=0.85]{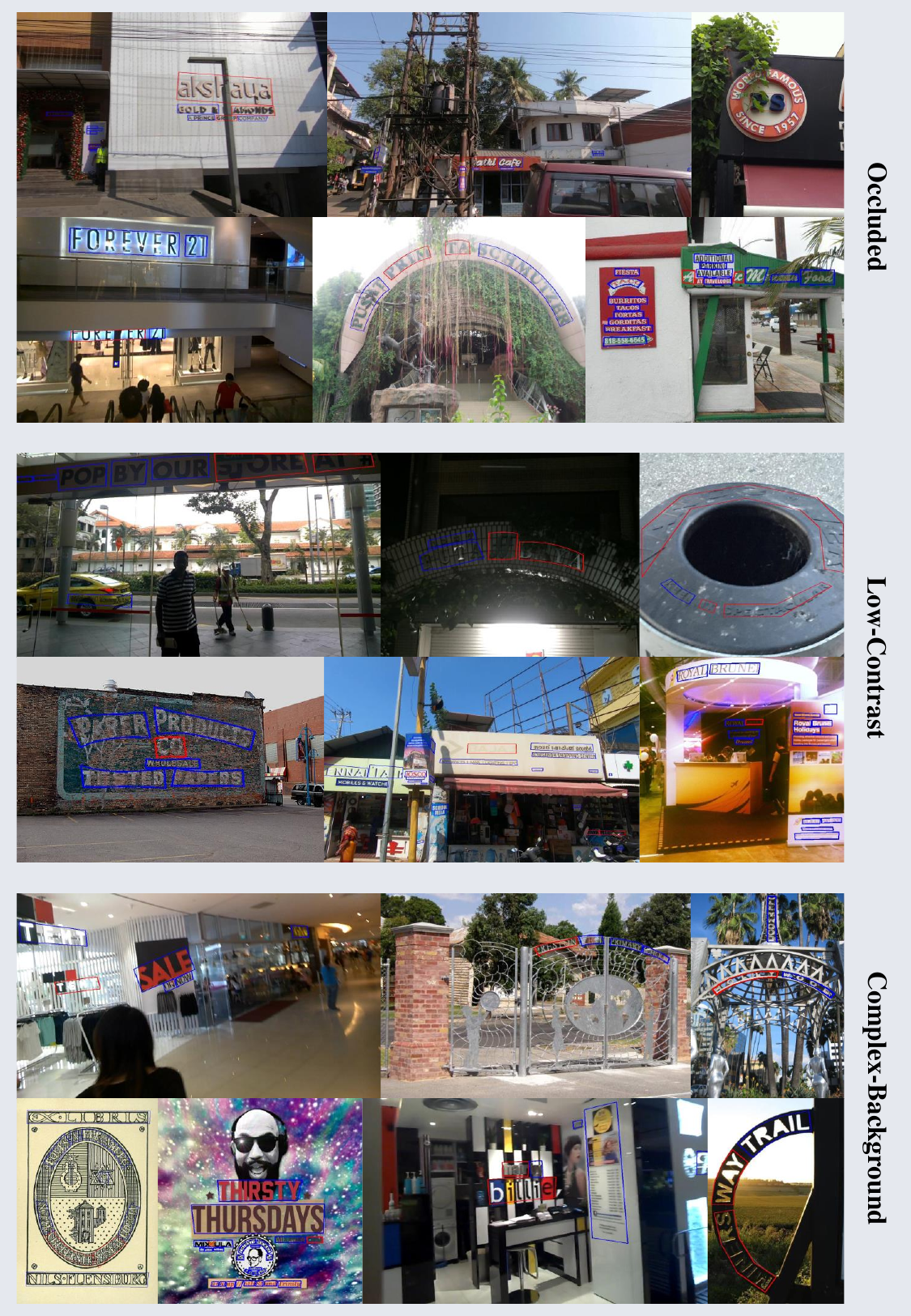}
  \caption{More examples about the background challenges of LTB. Red bounding boxes mark challenging text instances and blue ones for \textit{don't care} regions.
  }
  \label{fig:more_ltb_examples_4}
\end{figure*}

\end{document}